\documentclass[journal]{IEEEtai}

\usepackage[colorlinks,urlcolor=blue,linkcolor=blue,citecolor=blue]{hyperref}

\usepackage{color,array}

\usepackage[numbers]{natbib}
\usepackage{ragged2e}
\usepackage{amsmath}
\usepackage{amssymb}
\usepackage{hyperref}
\usepackage{booktabs}
\usepackage[ruled,linesnumbered]{algorithm2e}
\usepackage{graphicx}
\usepackage{subfigure}
\usepackage{algpseudocode}
\usepackage{authblk}
\usepackage{multirow}
\usepackage{xspace}  
\newcommand{\eg}{e.g.\@\xspace}
\newcommand{\etal}{et~al.\@\xspace}
\newcommand{\ie}{i.e.\@\xspace}

\usepackage{xspace}
\newcommand{\vs}{vs.\@\xspace}

\begin{document}

\title{SGLP: A Similarity Guided Fast Layer Partition Pruning for Compressing Large Deep Models} 

\author[1]{Yuqi Li$\dagger$}
\author[2]{Yao Lu$\dagger$\thanks{$\dagger$: Equal contributions}}
\author[3]{Junhao Dong}
\author[1]{Zeyu Dong}
\author[1]{Chuanguang Yang}
\author[4]{Xin Yin}
\author[4]{Yihao Chen}
\author[5]{Jianping Gou*\thanks{Corresponding author: pjgzy61@swu.edu.cn}}
\author[6]{Yingli Tian, \textit{IEEE Fellow}}
\author[7]{Tingwen Huang, \textit{IEEE Fellow}}
\affil[1]{Institute of Computing Technology, Chinese Academy of Sciences, China}
\affil[2]{Zhejiang University of Technology, China}
\affil[3]{Nanyang Technological University, Singapore}
\affil[4]{Zhejiang University, China}
\affil[5]{Southwest University, China}
\affil[6]{The City College of New York, New York, USA}\affil[7]{Shenzhen University of Advanced Technology, Shenzhen, China}

\maketitle

\begin{abstract}
The rapid advancement of Deep Neural Network (DNN)-based networks has revolutionized deep learning. However, the deployment of these networks on resource-constrained devices remains a significant challenge due to their high computational and parameter requirements. To solve this problem, layer pruning has emerged as a potent approach to remove redundant layers in the pre-trained network on the purpose of reducing network size and improve computational efficiency. However, existing layer pruning methods mostly overlook the intrinsic connections and inter-dependencies between different layers within complicated deep neural networks. This oversight can result in pruned models that do not preserve the essential characteristics of the pre-trained network as effectively as desired. To address these limitations, we propose a \textbf{S}imilarity-\textbf{G}uided \textbf{L}ayer \textbf{P}artition (\textbf{\textit{SGLP}}) Pruning, a novel pruning framework that exploits representation similarity to guide efficient and informed layer removal for compressing large deep models.  Our method begins by employing Centered Kernel Alignment (CKA) to quantify representational similarity between layers, uncovering structural patterns within the network. We then apply Fisher Optimal Segmentation on the similarity matrix to partition the network into semantically coherent layer segments. This segmentation allows pruning decisions to respect layer interdependencies and preserve essential knowledge. Within each segment, we introduce a fine-tuning-free importance evaluation using GradNorm, identifying and removing redundant layers in a targeted, segment-wise manner. 
Experimental results on both image classification tasks and large language models (LLMs) demonstrate that our proposed \textit{SGLP} outperforms the state-of-the-art methods in accuracy and efficiency. Our approach achieves significant model compression with minimal performance degradation, making it well-suited for deployment in resource-limited environments. 
\end{abstract}




\section{Introduction}

Deep neural networks (DNNs) have demonstrated unparalleled effectiveness in various computer vision tasks \cite{resnet,ou2025analyzing,ou2025social}. Modern DNNs have achieved unprecedented success, particularly in computer vision tasks. However, these high-performing networks often come with a significant computational and parameter burden, posing a considerable bottleneck for deployment on resource-constrained devices such as mobile and embedded systems. Existing network compression methods, including weight pruning and filter pruning, have somewhat alleviated this issue but still fall short of achieving substantial compression while maintaining network performance. Even with innovations like efficient architectural designs, \eg residual connections ~\cite{resnet}, the issue of over-parametrization still exists. To tackle this problem, it is necessary to obtain a network with smaller memory footprints and computational costs while maintaining its accuracy.

Network pruning is a commonly-used way to compress a neural network, which can mainly categorized as unstructured pruning and structured pruning. Unstructured pruning, such as weight pruning, removes the unimportant weights from the network, resulting in a sparse tensor. However, networks derived from unstructured pruning can only be integrated into BLAS libraries and requires specialized hardware for acceleration, which hinder the practical usage. In contrast, structured pruning discards entire structures in a network, resulting in a pruned network which does not rely on specialized software or hardware. Among structured pruning, filter/channel pruning compresses neural networks by removing unimportant filters/channels in CNN-based networks \cite{pfec, fpgm}. For example, Lin \etal \cite{hrank} remove the filters with the least rank. However, filter/channel pruning faces constraints caused by the depth of the original network, as each layer is required to retain a minimum of one filter or channel. Besides, the parallel nature of convolutional operations across various filters within the same layer means that channel pruning may not result in significant performance gains on hardware that supports parallel processing. Additionally, the impact on reducing memory usage during runtime may be limited. Therefore, such methods do not yield significant acceleration on hardware platforms and do not notably reduce runtime memory overhead. 

On the contrary, layer pruning~\cite{li2025sepprune} discards entire redundant layers and reduces the depth of a network without disrupting its structure, making it a more fitted choice for deployment on resource-constrained devices \cite{sr-init,lu2024reassessing}. For instance, when pruning the layers, the work \cite{representations} leverages linear classifier probes as the importance criterion of the layers and removes the least important ones. In addition, Ke \etal \cite{shallower} propose to disconnect the less significant residual mappings with Taylor expansion, retaining only the identity mapping. This approach effectively removes the corresponding residual module and accomplishes the goal of layer pruning. 

While existing pruning methods have demonstrated notable success in reducing the complexity of deep neural networks, several limitations have been identified that our new method aims to address. Traditional pruning techniques often fail to account for the intrinsic connections and inter-dependencies between layers within complex networks, which can lead to pruned models that are less capable of preserving the essential features of the original models. This oversight may result in a loss of critical information and a degradation in performance post-pruning. To tackle these limitations, we focus on layer pruning while leveraging the internal representation similarities of the pre-trained network, aiming at removing parameters and reducing computational overhead while maintaining the performance of the network. Specifically, we propose a \textbf{S}imilarity \textbf{G}uided fast \textbf{L}ayer \textbf{P}artition pruning for compressing large deep models (\textbf{\textit{SGLP}}), anchoring in the observation that traditional pruning approaches either ignore the intrinsic relationships between layers or require extensive computation to identify redundant layers. The core idea behind \textit{SGLP} is to exploit the layer-wise representation similarities as a guidance for efficient network segmentation, thereby identifying and removing redundant layers in a more targeted and informed manner. Our method is underpinned by the belief that a well-segmented network can reveal layers that contribute less to the overall performance, allowing for more precise pruning.

Our method introduces a three-step process that begins with innovatively applying Centered Kernel Alignment (CKA) \cite{cka} to identifying redundant layers within deep neural networks by quantitatively measuring the representation similarity between layers. By calculating the similarity matrix between layers, we gain a comprehensive understanding of the redundancy across the network. Traditional layer pruning methods often evaluate each layer's importance individually, which is computationally expensive and overlooks the collective impact of layers. In contrast, we employ Fisher Optimal Segmentation \cite{fisher-segmentation} to partition the model into blocks based on the similarity matrix derived from CKA. When pruning, we identify the redundant layers using GradNorm \cite{gradnorm} to evaluate the performance of layer combinations within each segment and prune the unimportant ones to obtain a compact network. Finally, we fine-tune the pruned network to recover its performance. Extensive experiments in image classification and for large language models (LLMs) show that our \textit{SGLP} significantly outperforms other state-of-the-art layer pruning methods in accuracy and computational efficiency, which demonstrates the effectiveness and superiority of our proposed method.

\textbf{Contributions: } The contributions of our proposed layer pruning method \textit{SGLP} are summed as follows:
\begin{enumerate}
\item Similarity-Guided Layer Partitioning:  We introduce a novel strategy for layer pruning by integrating internal representation similarity with optimal network segmentation. Using CKA, we construct a similarity matrix that captures the relationships between layers. Fisher Optimal Segmentation is then applied to divide the network into segments with coherent representational characteristics. This segment-wise approach preserves critical network functionality and mitigates the risk of excessive information loss during pruning.
	
\item Efficient Segment-Wise Pruning via GradNorm: Rather than evaluating layer importance across the entire network, we focus on intra-segment redundancy. Within each segment, we apply GradNorm to assess the contribution of each layer without requiring fine-tuning. This targeted strategy significantly reduces computational overhead while maintaining high pruning precision.

\item Strong Performance Across Domains: Our method is extensively validated on both image classification and LLMs. SGLP outperforms existing pruning techniques in terms of both accuracy retention and computational efficiency. Ablation studies further demonstrate that our similarity-guided segmentation reduces the loss of critical information typically associated with aggressive layer pruning.
	
\end{enumerate}

\section{Related Work}

\subsection{Network Pruning}

Network pruning is an effective approach aiming at reducing the size of neural networks by eliminating unnecessary patterns. It is commonly divided by three categories, which remove the weights, filters/channels or layers of neural networks, respectively.

\textbf{\textit{(1) Weight Pruning: }} In the early work by LeCun \etal \cite{obd}, they explore the importance of weights in neural networks and introduce a pruning method based on weight gradients. Recent work \cite{comparing} measure the importance of each neuron based on the Hessian matrix of the loss function, thereby reducing the number of connections in the network. Experimental validation of these methods has shown that pruning based on the Hessian matrix can achieve better network performance compared to pruning based on the magnitude of weights. However, unstructured weight pruning faces a limitation that achieving speed-up from compression requires dedicated hardware and libraries. The recent study \cite{fontana2024distilled} presents Distilled Gradual Pruning with Pruned Fine-tuning (DG2PF) to improve the efficiency of neural networks through a combination of magnitude-based unstructured pruning and knowledge distillation. 

\textbf{\textit{(2) Filter/Channel Pruning: }} Filter/channel pruning methods discard filters or channels in the neural networks, respectively, and they are closely related. Previous studies \cite{pfec} employs $\ell_1$-norm as importance criteria to discard the unimportant filters for network compression. FPGM \cite{fpgm} first calculates the geometric median for each convolutional layer, and then assumes that convolutional filters close to the geometric median can be replaced by other filters. Therefore, pruning these filters close to the geometric median is expected to have no substantial negative impact on the performance of the network. HRank \cite{hrank} first mathematically proves that convolutional filters corresponding to feature maps with relatively small rank contain less information. Therefore, their ability to maintain network accuracy is relatively weak. During pruning, the priority is given to removing convolutional filters with smaller rank, thus achieving network compression. Recent research \cite{cicc} utilizes Shapley values as importance criteria to remove unimportant channels. UDSP \cite{gao2024bilevelpruning} proposes a novel bi-level optimization-based model that integrates both dynamic and static channel pruning for convolutional neural networks, optimizing the static sub-network through dynamic sub-networks. Besides, RLAL \cite{ganjdanesh2024jointly} proposes a novel method for jointly training and pruning Convolutional Neural Networks (CNNs) using a reinforcement learning (RL) agent. The RL agent determines the pruning ratios of the CNN layers, with the model's accuracy serving as the reward, and a recurrent model is employed to represent the dynamic environment for the agent to effectively learn a proper policy. AFIE \cite{lu2024entropy} decomposes the weight matrix of each layer into a low-rank space, quantifies filter importance based on the distribution of normalized eigenvalues, and assigns pruning ratios accordingly.

Nevertheless, these pruning methods which compress neural networks at the channel or filter level pose challenges in attaining a substantial compression rate and practical acceleration.

\textbf{\textit{(3) Layer Pruning: }} Layer pruning, as opposed to filter/channel pruning that narrows the network width, removes entire layers of neural networks, thereby making the networks shallower. Unlike filter/channel pruning, layer pruning yields a more substantial reduction in run-time memory. Additionally, as convolutional computations on the same layer often occur in parallel, discarding convolutional layers can result in greater speed-up. Consequently, the practical acceleration achieved by layer pruning outperforms that of filter/channel pruning, especially on edge devices. Recent work \cite{representations} introduces a layer pruning approach where layers with less improvements in feature representations are pruned. Previous study \cite{discriminative} prunes the layers according to partial least squares projection as a discriminative importance estimation criterion. Elkerdawy \etal \cite{question} propose a framework that removes the layers of the network with low importance scores derived from imprinting, after which they fine-tune the compact network to recover the performance. Recent work GAL \cite{gal} employs GAN (generative adversarial network) for pruning the layers as well as filters simultaneously. In addition, the previous work \cite{elkerdawy2020one} introduces an one-shot layer-wise proxy classifier method for layer pruning in neural networks, which estimates the importance of each layer to prune entire layers and accelerate inference speed. The method \cite{hossain2024novel} leverages global standard deviation pooling and efficient attention mechanisms to identify and prune the least significant layers.

Compared to existing methods, the uniqueness of our proposed \textit{SGLP} lies in its deep understanding and utilization of the relationships between layers. While traditional pruning methods often independently assess the importance of each layer, our method identifies layers that can be safely removed by analyzing the similarities between layers. Furthermore, GradNorm used in our presented \textit{SGLP} assesses layer importance without the need for time-consuming fine-tuning, significantly accelerating the pruning process and reducing the demand for computational resources.

\begin{figure*}[t]
	\centering
	
	\includegraphics[width=\linewidth]{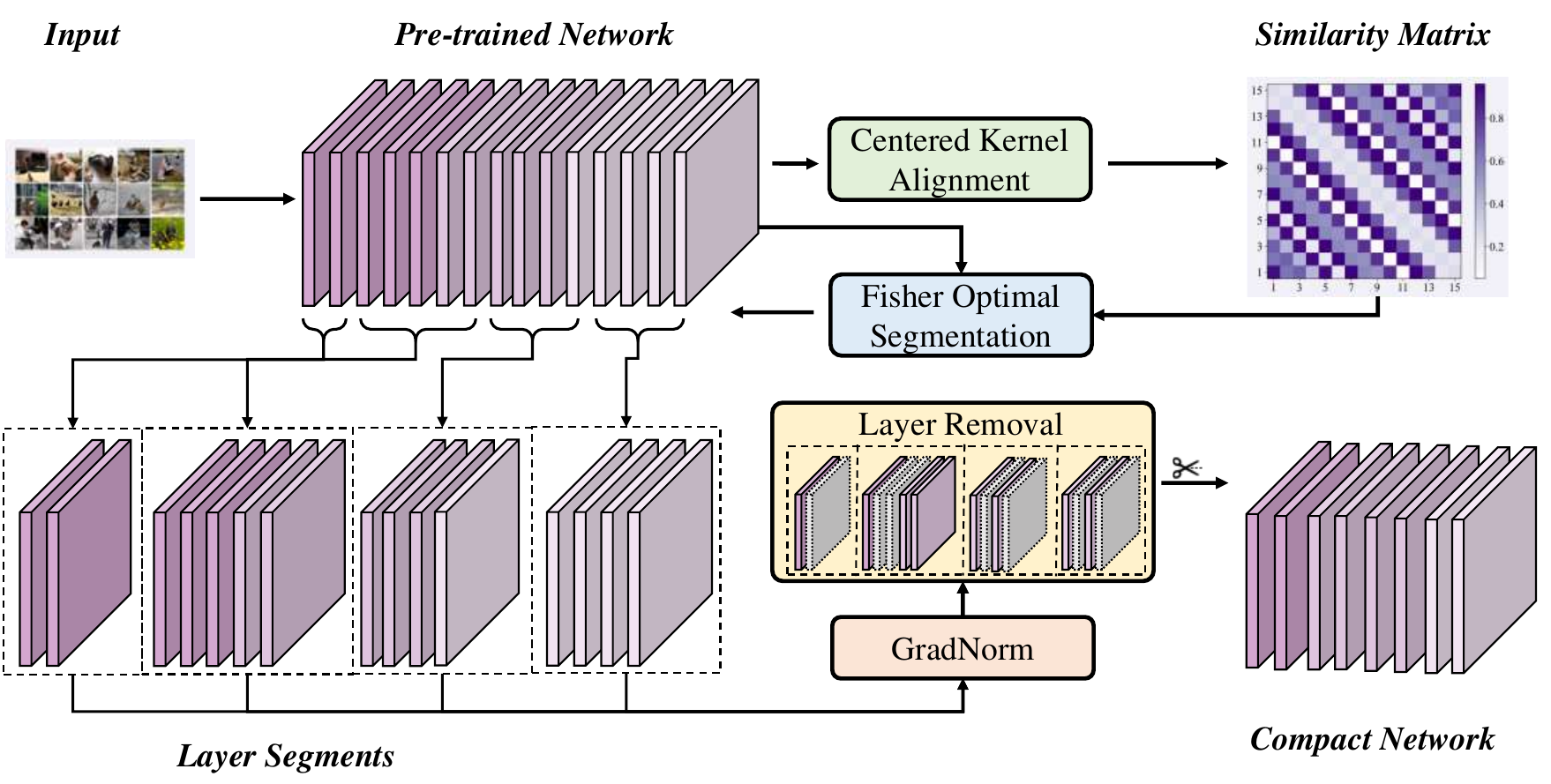}
	
	\caption{The overview framework of our proposed method. We take a 15-layer network as an example, where the squares in gray denote the layers with the lowest importance, which will be discarded when pruning. We first feed batches of inputs to the pre-trained network for forward propagation. Then, the similarity matrix is derived via Centered Kernel Alignment, which indicates the representations similarity among the layers. Based on the layer similarity, we partition the network into layer segments via Fisher Optimal Segmentation, which provides a basis for subsequent layer pruning. In each layer segment, we evaluate the importance for the layers via GradNorm, where the unimportant ones are removed to obtain a compact network.}
	\label{fig:framework}
\end{figure*}

\subsection{Representation Similarity}

The similarity of the neural network representations is commonly used to understand and characterize the representations of neural networks learned from the given data. For example, Content and Cluster Analysis \cite{content} employs the distances between neural activations for comparing neural representations, and empirically demonstrates that different networks which are trained through back-propagation on identical categorization tasks can converge to the states where the hidden representations are similar. Canonical Correlation Analysis (CCA) \cite{cca} identifies bases for two matrices, and the correlation is maximized when the original matrices are projected on them. SVCCA \cite{svcca} proposes singular vector canonical correlation analysis which is efficient to measure the intrinsic dimensionality and representations of different layers or networks. Previous study \cite{extent} formulates a robust theory based on the subspace matching network of neuron activations. In addition, it proposes algorithms aiming to investigate the simple and maximum matches efficiently. Centered Kernel Alignment (CKA) \cite{cka} points out that CCA and its related methods which are invariant to invertible linear transformation are not able to assess meaningful similarities between higher-dimension representations. Hence, it proposes a similarity index that is equivalent to CKA and shares a close connection with CCA to estimate the relationship among representation similarity matrices, overcoming the limitations mentioned above.

Upon reviewing the existing network pruning methods, we have identified that existing methods mostly overlook the intrinsic connections and inter-dependencies between layers when dealing with complex deep neural networks, which may lead to pruned models that fail to effectively maintain the key features of the original models. To tackle this issue, the core of our proposed \textit{SGLP} lies in leveraging Centered Kernel Alignment (CKA) to capture the internal representation similarities among layers of a pre-trained network, and employing Fisher Optimal Segmentation to partition the network into multiple segments, which provides a basis for removing the layers in a segment-wise manner. Our presented method not only improves pruning efficiency but also preserves the accuracy of the network and generalization capability by more selectively removing redundant layers.

\section{Methodology}

\subsection{Motivation}

As deep learning has progressed, Deep Neural Networks (DNNs) have become the backbone of various tasks, particularly in areas such as image recognition and natural language processing. However, with the increasing depth and complexity of network architectures, the parameter count and computational demands of DNN models have surged, posing a significant challenge for deployment on resource-constrained devices, such as mobile and embedded systems. To address this issue, existing layer pruning methods have emerged to remove redundant layers within neural networks, on the purpose of obtaining a compact network. Nevertheless, they often face the following challenges:
\begin{enumerate}
	\item Traditional pruning methods may overlook the intrinsic connections and interdependencies between layers when assessing layer importance, which can result in pruned models that fail to effectively preserve the essential characteristics of the original models.
	\item The pruning decisions made by conventional methods may lack precision due to a lack of deep understanding of the functionality of each layer, leading to suboptimal pruning outcomes that may compromise the performance of the network.
	\item The process of identifying and removing unimportant layers through traditional methods is often computationally inefficient, requiring extensive fine-tuning, which can be both time-consuming and resource-intensive.
\end{enumerate} 

To address these challenges, as shown in Fig. \ref{fig:framework}, we propose a \textbf{S}imilarity \textbf{G}uided fast \textbf{L}ayer \textbf{P}artition pruning for compressing large deep models (\textbf{\textit{SGLP}}). Specifically, our proposed method contains a three-step pruning framework that begins with Centered Kernel Alignment (CKA) \cite{cka} to quantitatively measure the representation similarity between layers of the pre-trained network. Then, we employ Fisher Optimal Segmentation \cite{fisher-segmentation} which leverages the CKA-derived similarity matrix to partition the network layers into segments, providing a basis for removing the layers in a segment-wise manner. When pruning, we identify the redundant layers using GradNorm \cite{gradnorm} to evaluate the performance of layer combinations within each segment and prune the unimportant ones to obtain a compact network.

In summary, the key advantages of our method are as follows:
\begin{enumerate}
	\item Our proposed approach quantifies and leverages the representational similarities between layers using Centered Kernel Alignment (CKA), which aids in identifying and removing redundant layers that exhibit high overlap in their feature representations.
	\item We employ Fisher Optimal Segmentation to effectively partition the network into segments, which provides a basis for removing the layers in a segment-wise manner.
	\item We also incorporate GradNorm for an efficient evaluation of layer importance, eliminating the need for extensive fine-tuning and enabling rapid and accurate identification of dispensable layers.
\end{enumerate}

\subsection{Problem Formulation}
\label{sec:pre}

We suppose that a pre-trained network $\mathcal{N}$ with weights $\mathcal{W}$ contains $L$ layers $\mathbb{L} = \{l_1, l_2, \ldots, l_L\}$ trained on the data $x = \{x_1, x_2, \ldots, x_N\}$ with labels $y = \{y_1, y_2, \ldots, y_N\}$, where $N$ represents the size of the dataset. In layer pruning, $L$ layers in $\mathcal{N}$ are divided into two groups, \ie, the removed unimportant subset $\mathcal{U} = \{l_{\mathcal{U}_1}, l_{\mathcal{U}_2}, \ldots, l_{\mathcal{U}_{n_1}}\}$ and the kept important subset $\mathcal{I} = \{l_{\mathcal{I}_1}, l_{\mathcal{I}_2}, \ldots, l_{\mathcal{I}_{n_2}}\}$, where $n_1$ and $n_2$ denote the number of unimportant and important layers, respectively. Here, we have:
\begin{equation}  
	\left\{  
	\begin{array}{lr}  
		\mathcal{U} \cup \mathcal{I} = \mathbb{L}, &  \\  
		\mathcal{U} \cap \mathcal{I} = \varnothing, & \\
		n_1 + n_2 = L.
	\end{array}  
	\right.  
\end{equation}

Assume that $\phi (l_i)$ denotes the importance of the $i_{th}$ layer in $\mathcal{N}$, thus layer pruning can be represented as an optimization problem:
\begin{equation}
		\min\limits_{\delta_i} \sum\limits_{i=1}^{L} \delta_i \phi(l_i), \quad
		s.t. \; \sum_{i=1}^{L} \delta_i = n_1,
\end{equation}
where $\delta_i$ is an indicator which equals to $1$ when $l_i \in \mathcal{U}$ or $0$ when $l_i \in \mathcal{I}$. The objective of layer pruning is to minimize the information of the discarded layers.

\subsection{Layer Representation Similarity}
\label{sec:cka}

Traditional pruning methods, while effective in reducing the size and computational demands of neural networks, often employ strategies that consider each layer in isolation when evaluating its importance to the  overall performance of the network. Such approaches may overlook the complex interactions between layers, where the collective behavior of layers can be more indicative of their true contribution to the functionality of the network than individual assessments. To address these limitations, a more sophisticated pruning methodology is necessary, \ie, one that takes into account the intrinsic connections and inter-dependencies between layers and evaluates the importance of layers within the context of the network as a whole. With this strategy, we can ensure that the pruning process not only leads to a more efficient network but also one that preserves the fundamental qualities that make the original model effective in the first place.

To this end, considering that Centered Kernel Alignment (CKA) \cite{cka} has great potential to capture more meaningful similarities since it is not invariant to all invertible linear transformations \cite{cca, svcca}, we employ it to measure the representational similarities between layers of a pre-trained network, which aids in understanding the intrinsic connections and inter-dependencies among different layers. CKA is a statistical measure that captures the similarity between the representations learned by different layers in a neural network, which provides us with a potent basis for layer pruning, where the derived similarity matrix reflects the relationships between all pairs of layers. In addition, by focusing on the representation similarity, we can ensure that the pruned network maintains the qualitative aspects of its feature representations, which are critical for preserving the performance of the network.

The similarity analysis process commences with organizing the data into a matrix, where each entry corresponds to a feature of a data point. A distance metric is then applied to calculate the pairwise dissimilarities between all data points, populating a symmetric dissimilarity matrix. To enhance efficiency, only the upper triangular part of this matrix is stored, which is later used to reconstruct the full matrix if needed. The final dissimilarity matrix is analyzed to uncover patterns and relationships within the data, providing valuable insights into the underlying structure.

Thus, the application of CKA serves as the cornerstone of our approach, providing a data-driven basis for layer analysis and selection. Given a pre-trained network with \( L \) layers, we first feed a batch of data \( \mathcal{X} = \{x_1, x_2, \dots, x_B\} \) through the network to obtain the activations for each layer, where \( B \) is the batch size. Let \( z_i \in \mathbb{R}^{B \times C_i \times H_i \times W_i} \) denote the activation map of the \( i \)-th layer, where \( C_i \) is the number of channels, \( H_i \) is the height, and \( W_i \) is the width of the activation map.

To simplify the calculations, we apply a \textbf{flattening operation} to each activation map \( z_i \), transforming it into a matrix \( \hat{z}_i \in \mathbb{R}^{B \times (C_i H_i W_i)} \). This flattening operation combines the spatial dimensions into a single dimension, preserving the number of samples \( B \) while collapsing the remaining dimensions into one.

Next, we compute the Gram matrices for each layer. The Gram matrix \( K_i \) for layer \( i \) is defined as:
\begin{equation}
    K_i = \hat{z}_i \hat{z}_i^\top \in \mathbb{R}^{B \times B}.
\end{equation}

To center the Gram matrices, we use the centering matrix \( H \) defined as:
\begin{equation}
    H = I_B - \frac{1}{B} \mathbf{1} \mathbf{1}^\top,
\end{equation}
where \( I_B \) is the \( B \)-dimensional identity matrix, and \( \mathbf{1} \) is a \( B \)-dimensional column vector of all ones.

The centered Gram matrix \( K_i^c \) is then computed as:
\begin{equation}
    K_i^c = H K_i H.
\end{equation}

The traces of the original and centered Gram matrices are calculated as:
\begin{equation}
    \label{eq:trace_1}
    \text{trace}(K_i) = \sum_{b=1}^B K_i(b, b),
\end{equation}
\begin{equation}
    \text{trace}(K_i^c) = \sum_{b=1}^B K_i^c(b, b).
\end{equation}

The squared Frobenius norms of the original and centered Gram matrices are calculated as:
\begin{equation}
    \|K_i\|_F^2 = \text{trace}(K_i K_i^\top),
\end{equation}
\begin{equation}
    \label{eq:trace_2}
    \|K_i^c\|_F^2 = \text{trace}(K_i^c K_i^{c\top}).
\end{equation}

Finally, the CKA similarity between layers \( i \) and \( j \) is computed as:
\begin{equation}
    \text{CKA}(K_i, K_j) = \frac{\text{trace}(K_i^c K_j^c)}{\|K_i^c\|_F \|K_j^c\|_F}.
\end{equation}

The $2$-dimension similarity matrix \( M \in \mathbb{R}^{L \times L} \) of $L$ rows and $L$ column is then constructed, where each element \( M_{ij} \) represents the CKA similarity between layers \( i \) and \( j \):
\begin{equation}
    \label{eq:similarity}
    M_{ij} = \text{CKA}(K_i, K_j).
\end{equation}

This similarity matrix provides a quantitative measure of the representational similarity between all pairs of layers in the network, which serves as the basis for our layer pruning strategy.

\subsection{Network Partitioning}
\label{sec:fisher}

Existing layer pruning methods usually evaluate the importance of each layer separately, which is time-consuming in the networks consisting of numerous layers. Thus, it is necessary to partition the network into multiple segments, based on which we can focus on the layers within these segments rather than conducting importance evaluation through all layers of the whole network, thereby reducing the computational cost and improving the efficiency of the layer pruning process.

In ordered sample clustering, it is required that the samples be arranged in a certain order, and the order should not be disrupted during classification, meaning that samples of the same category must be adjacent to each other. That is, if $\{c_1, c_2, \cdots, c_m\}$ denotes $m$ ordered samples, then each cluster must follow $\{c_i, c_{i+1}, \cdots, c_{i+h}\}$, where $1 \leq i \leq m, h \geq 0$ and $i + h \leq m$. In other words, the samples in a cluster must be adjacent to each other. Clustering of ordered samples essentially identifies certain partition points that divide the ordered samples into several segments, with each segment being considered a separate category. A common criterion for finding the best segmentation is to minimize the differences among samples within each segment while to maximize the differences between samples from different segments. To this end, we employ Fisher Optimal Segmentation to accomplish these objectives, which divides data according to their natural order, leading to maximizing the intra-segment similarity and minimizing the inter-segment difference among the categories while ensuring that the segments are coherent and distinct from one another \cite{fisher-segmentation}. 

In our layer pruning scenario, considering that each layer in a neural network also has a specific order, the layers in the network can exactly be viewed as sequentially ordered data. Hence, we extend Fisher Optimal Segmentation to layer pruning to partition the layers of a neural network while distinguishing the layers effectively. When segmenting layers in a neural network, maximizing the similarity within each segment ensures that the network can retain sufficient feature representation capabilities within each segment even after pruning. This helps to maintain the integrity of the network performance, since each segment can continue to capture the essential characteristics of the data it processes. On the other hand, maximizing the differences between segments provides a stronger degree of discrimination for the layers. 

In addition, as introduced in Sec. \ref{sec:cka}, CKA quantifies the similarity of representations between neural network layers, providing a deeper understanding of the internal workings of the network. Therefore, we implement Fisher Optimal Segmentation based on the similarity matrix derived from CKA, ensuring that the partitioning process takes the intrinsic connections and representation similarity among the layers into account with more adequate neural interpretability. Thus, this combination enables us to preserve significant information of the network, contributing to less loss resulted from layer pruning.

The core of Fisher Optimal Segmentation involves finding the optimal partition of the data into segments such that the intra-segment similarity is maximized and the inter-segment difference is minimized. This is achieved by defining a loss function that calculates the sum of the diameters for all potential segments. The optimal partition is determined by recursively partitioning the data into segments that minimize this loss function, leading to the most distinct and coherent segments.

Let $A_i$ denote the sum of the elements in each row:
\begin{equation}
	\label{eq:sum}
	A_i = \sum_{j=1}^L M_{i,j}.
\end{equation}
Thus, $A_i$ is a vector of $L$ elements.

Since the order of $L$ layers cannot be variant, the matrix $A = \{A_i\} (1 \leq i \leq L)$ can be regarded as an ordered sequence containing $L$ samples. In ordered clustering, we suppose that $P(L, k)$ denotes $k$ segments of $L$ ordered samples, where $k \leq L$. Then, the segmentation result should maintain the original sequence of the data, which can be denoted as:
\begin{equation}
	\begin{aligned}
		& \{A_{i_1}, A_{i_1+1}, \cdots, A_{i_2-1}\}, \\ 
		& \{A_{i_2}, A_{i_2+1}, \cdots, A_{i_3-1}\}, \\ 
		& \cdots, \\
		& \{A_{i_k}, A_{i_k+1}, \cdots, A_{L}\},
	\end{aligned}
\end{equation}
where $1 = i_1 \textless i_2 \cdots i_k \leq L$.

The total number of segmentation results for dividing $L$ ordered samples into $k$ segments is:
\begin{equation}
	\label{eq:total_k}
	C_{L-1}^{k-1} = \frac{(L-1)!}{(k-1)!(L-k)!},
\end{equation}
where $C$ refers to the selection of items from a larger set, where the order of the items does not matter.

The result in Eqn. (\ref{eq:total_k}) is under the premise of the fact that the number of segments $k$ is specified. Obviously, $L$ ordered data can be divided into $2, \cdots, k$ segments, where $k \leq L$. Thus, the number of possible segments of $L$ ordered data is:
\begin{equation}
	\label{eq:total_p}
	P(L, (2, \cdots, k)) = C_{L-1}^{1} + C_{L-1}^{2} + \cdots + C_{L-1}^{k-1} = 2^L-1.
\end{equation}

Among the $2^L-1$ segmentation results derived from Eqn. (\ref{eq:total_p}), there must be one or more segmentations that lead to the smallest intra-segment difference and the largest inter-segment difference. To this end, we employ Fisher Optimal Segmentation \cite{fisher-segmentation} on the purpose of deriving such optimal segmentations.

The sum of squared deviation is a concept in statistics which refers to the sum of the squared differences between each item and the average value of the given data, measuring how discrete a set of data. Hence, we use $A_a$ to denote the sum of squared deviation of the similarity matrix:
\begin{equation}
	\label{eq:squared}
	A_a = \sum_{i=1}^L (A_i - \overline{A})^2,
\end{equation}
where $\overline{A}$ represents the average value of $A$. The larger the sum of squared deviation, the greater the degree of dispersion of the data.

Fisher Optimal Segmentation calculates the sum of squared deviations for each of the $C_{L-1}^{k-1}$ possible segments given the number of segments $k$. It takes the smallest segment as the optimal segmentation under the number of segments $k$.


Let a segmented section be $\{A_{r}, A_{r+1}, \cdots, A_{s}\}$, where $s \textgreater r$. Thus, we employ ``diameter'' $D(r, s)$ to estimate the difference between the similarity of layers:
\begin{equation}
	D(r, s) = \sum_{i=r}^s (A_i - \overline{A})^2.
\end{equation}

Hence, given $L$ and $k$, the optimal segmentation is denoted as below:
\begin{equation}
	\mathcal{L}^*(L, k) = \mathop{\min}_{1 \leq i \leq L} \sum_{l=1}^k D(i_l, i_l-1),
\end{equation}
where $\mathcal{L}^*$ denotes the classification loss function.

The optimal segmentation of an ordered sequence is formed by adding a segment after the optimal ${k-1}_{th}$ segmentation of a truncated segment $P^*(i, k - 1)$. Therefore, there are two recurrence formulas as follows:
\begin{equation}
	\label{eq:recurrence_1}
	\mathcal{L}^*(L, 2) = \mathop{\min}_{2 \leq i \leq L} \{D(1, i-1)+D(i, L)\}.
\end{equation}
\begin{equation}
	\label{eq:recurrence_2}
	\mathcal{L}^*(L, k) = \mathop{\min}_{k \leq i \leq L} \{\mathcal{L}^*(i-1, k-1)+D(i,L)\}.
\end{equation}

According to Eqn. (\ref{eq:recurrence_1}) and Eqn. (\ref{eq:recurrence_2}), if $L$ and $k$ are both specified, the optimal segmentation point $i_k$ can be determined when the loss function reaches its minimum:
\begin{equation}
	\mathcal{L}^*(L, k) = \mathcal{L}^*(i_k - 1, k - 1) + D(i_k, L).
\end{equation}

Then, continue searching for the split point $i_{k - 1}$ so that the loss function $\mathcal{L}^*(i_k - 1, k - 1)$ is minimized:
\begin{equation}
	\label{eq:fos_loss}
	\mathcal{L}^*(i_k - 1, k - 1) = \mathcal{L}^*(i_{k - 1} - 1, k - 2) + D(i_{k - 1} - 1, k - 1).
\end{equation}


Above all, once $k$ is decided, the optimal segmentation for $L$ layers will be derived. Thus, $k$ is a critical hyper-parameter which requires our heuristic tuning. Once the loss function Eqn. \ref{eq:fos_loss} is minimized, the layer segments via Fisher Optimal Segmentation are derived, which yield the largest intra-segment similarity and inter-segment difference. By maximizing the similarity among the layers within a network segment, it is more possible to accurately identify the layers which have less impact on network performance. With layer segments partitioned by Fisher Optimal Segmentation, we can focus on each segment for layer pruning, instead of investigating the importance in a layer-wise manner. Thus, it effectively reduces the computation overhead in layer pruning.

\subsection{Layer Pruning}
\label{sec:gradnorm}

The key procedure of layer pruning is to evaluate the importance of the layers in a network, then identify and remove the unimportant ones. To this end, traditional pruning methods often rely on time-consuming fine-tuning to assess the impact of removing individual layers. For example, PFEC \cite{pfec} discards the layers via their sensitivity to the network. With such a sensitivity-based framework, it needs to first arrange and enumerate all the layers in the network, and initialize the selected layers with the weights of the pre-trained network, while the remaining layers are randomly initialized. Then, the performance of the network is derived via fine-tuning the network with new weights, and this process layer-wisely repeats to obtain the importance of each layer. Finally, the unimportant layers are discarded. However, such layer pruning process results in remarkable time overhead, since enumerating all the layers in the network leads to $2^L - 1$ combinations, and subsequent fine-tuning the network for each combination is time-consuming.

To solve this problem, efficient importance evaluation is of great necessity to be investigated, which simplifies and accelerates the process of layer pruning. GradNorm was proposed to balance the contribution of each task in a multi-task training procedure \cite{gradnorm}, while it is naturally fitted to evaluate the importance of layers efficiently. That is because that it aims to leverage the inherent information within the training dynamics of the network to guide our pruning decisions without any fine-tuning or training, providing a more efficient and accurate method for evaluating layer importance. Specifically, GradNorm operates by analyzing the gradients of a network during back-propagation, which provides a measure that how the parameters of each layer contribute to the outputs. By examining these gradients, the layers that have a minimal impact on the performance of the network can be identified, which indicates that they are less important and thus needs to be pruned. With GradNorm, we can quickly and accurately determine which layers are expendable, enabling the layer pruning process with high efficiency.

Sec. \ref{sec:fisher} illustrates that concentrating on the individual layers within the layer segments partitioned from the network, instead of performing an importance assessment across each layer in the network, can significantly reduce computational overhead in layer pruning. Thus, we implement importance evaluation for layers with GradNorm on the layer segments derived from Fisher Optimal Segmentation, instead of enumerating layers in the entire network. With Fisher Optimal Segmentation introduced in Sec. \ref{sec:fisher} based on Centered Kernel Alignment illustrated in Sec. \ref{sec:cka}, the neural network $\mathcal{N}$ is now partitioned to $k$ segments, which can be denoted as:
\begin{equation}
	Segment(\mathcal{N}, k) = \{ S_1, S_2, \ldots, S_k \},
\end{equation}
where $S_i$ represents the $i_{th}$ segment.

Then, we enumerate the combinations of the layers in each segment $S_i$ instead of all the layers in the network, yielding $2^{\lvert S_i \rvert} - 1$ combinations, where $\lvert S_i \rvert$ denotes the number of layers in $S_i$. Evaluating performance in a segment-wise manner leads to $k \cdot (2^{\lvert S_i \rvert} - 1)$ time complexity, which effectively decreases the number of combinations and thus increases the pruning efficiency.

Another critical issue is to save the time resulted from fine-tuning the pruned network which aims to derive the layer importance. Inspired by NASWOT \cite{naswot}, we present to prune the unimportant layers via GradNorm \cite{gradnorm} within each segment, which evaluates the performance according to the gradients of networks without time-consuming fine-tuning. Specifically, for the enumerated layers in $S_i$, we initialize the weights of them and others with the weights of the pre-trained network and random weights, respectively. Then, the network is back-propagated on the training data $x$ for generating gradients. Instead of fine-tuning the pruned network to investigate the performance of layers, we adopt the Euclidean-norm of gradients of all the parameters as the representation for the importance of the selected layers in $S_i$, which is denoted as:
\begin{equation}
	\label{eq:gradnorm}
	G_\mathcal{W} = \sqrt{\sum_{i=1}^{t} \| \nabla_{\mathcal{W}_i} \psi(\mathcal{W}_i; x, y) \|^2},
\end{equation}
where $t$ and $\mathcal{W}_i$ denote the total number of parameters and the $i_{th}$ weight of the network, respectively, and $\nabla_{\mathcal{W}_i} \psi$ represents the gradient of the optimization loss function $\psi$ with respect to the $i_{th}$ parameter $\mathcal{W}_i$.

The larger $G_\mathcal{W}$, the better the performance of the selected layers. Therefore, the layer segment $S_i$ becomes a combination of layers which achieve the maximum GradNorm:
\begin{equation}
	\label{eq:final}
	S_i^{\prime} = {\operatorname{arg\,max}}\, G_\mathcal{W}.
\end{equation} 

Thus, the corresponding layers which lead to the best performance in a segment are kept, while the remaining ones in the segment are pruned. Similar to the previous work \cite{representations}, for ResNet, we regard each residual block as an independent layer. 

In summary, Alg. \ref{alg:gradnorm} illustrates the entire pipeline our proposed layer pruning method. We first utilize Centered Kernel Alignment (CKA) to quantitatively measure the internal representational similarities among layers of the pre-trained network, generating a similarity matrix that serves as a potent basis for layer pruning. Then we apply Fisher Optimal Segmentation based on the CKA-derived similarity matrix to partition the network into multiple segments, ensuring high intra-segment similarity and distinct inter-segment differences. Within each layer segment, we employ GradNorm as an efficient criterion to assess the importance of layers, eliminating the need for extensive fine-tuning. Finally, we preserve the layers which achieve the largest GradNorm, while remove other unimportant layers, so that we can obtain a compact network. Through this pipeline, our proposed \textit{SGLP} significantly reduces the computational and parameter requirements of DNNs while maintaining performance, effectively enhancing the deployment efficiency of DNNs on resource-limited platforms.

\begin{algorithm}[ht!]
	\KwIn{The training data $x$, labels $y$, pre-trained network $\mathcal{N}$ with weights $\mathcal{W}$, and the numner of layer segments $k$.}
	
	\tcp{Stage 1: Derive the layer representation similarity via Centered Kernel Alignment.}
	
	Forward-propagate $\mathcal{N}$ on $x$ to generate activation set $Z = \{z_i\} $ across all the layers.
	
	Consider $Z$ as the kernel matrix, and calculate the corresponding matrices by Eqn. (\ref{eq:trace_1}) $\sim$ Eqn. (\ref{eq:trace_2}).
	
	Compute the similarity matrix $M$ with Centered Kernel Alignment:
	$M = \frac{trace(Z_c)}{\left\| Z \right\|_F \cdot \left\| Z_c \right\|_F}$.
	
	\tcp{Stage 2: Partition the network into layer segments via Fisher Optimal Segmentation.}
	
	Compute $A_i$ representing the sum of the elements in each row: $A_i = \sum_{j=1}^L M_{i,j}$.
	
	Calculate $A_a$ which denotes the sum of squared deviation of the similarity matrix: $A_a = \sum_{i=1}^L (A_i - \overline{A})^2$.
	
	Employ diameter as the estimation of the difference between the similarity of layers: $D(r, s) = \sum_{i=r}^s (A_i - \overline{A})^2$.
	
	Minimize the loss function to accomplish Fisher Optimal Segmentation: $\mathcal{L}^*(i_k - 1, k - 1) = \mathcal{L}^*(i_{k - 1} - 1, k - 2) + D(i_{k - 1} - 1, k - 1)$.
	
	$\mathcal{N}$ is partitioned into $k$ layer segments: $Segment(\mathcal{N}, k) = \{ S_1, S_2, \ldots, S_k \}$.
	
	\tcp{Stage 3: Prune the layers via GradNorm in a segment-wise manner.}
	
	\For {$i = 1; i \leq k; i++$} {
		$size = |S_i|$ \tcp{The number of layers in the segment.}
		
		\For {\text{each binary number} $0 \le i \le 2^{\lvert S_i \rvert}$} {
			\parbox[t]{0.8\linewidth}{$layers = \{S_i[j] \; | \; j  \text{ corresponds to a}
\\\text{set bit in the binary representation of } i\}$ \\ 
\tcp{Enumerate layer combinations in the segment.}} 
			
			Initialize the weights of $layers$ with $\mathcal{W}$, and other layers are randomly initialized.
			
			Back-propogate $\mathcal{N}$ on $x$ for gradients and calculate GradNorm $G_\mathcal{W}$: $G_\mathcal{W} = \sqrt{\sum_{i=1}^{t} \| \nabla_{\mathcal{W}_i} \psi(\mathcal{W}_i; x, y) \|^2}$.
		}
		
		\tcp{Remove the unimportant layers.}
		
		Preserve the layers in the segment which generate the largest GradNorm: $S_i^{\prime} = {\operatorname{arg\,max}}\, G_\mathcal{W}.$
	}
	
	\KwOut{The compact network $\mathcal{N}^{\prime}$.}
	
	\caption{Our proposed layer pruning pipeline.}
	
	\label{alg:gradnorm}
\end{algorithm}

The network partitioning is crucial for the subsequent layer pruning process, where unimportant layers are identified and removed based on their contribution to the network's performance, as assessed by GradNorm. By focusing on these segments instead of the entire network, the RSNP framework significantly reduces the computational overhead associated with layer pruning, leading to a more efficient and effective model compression technique.

\section{Experiments}

\subsection{Experimental Settings}

\subsubsection{Backbones and Datasets}

We conduct experiments for pruning VGGNet and ResNet on CIFAR-10, CIFAR-100, ImageNet, Imagenette2 and Imagewoof2. CIFAR-10 contains 60,000 images with 32 $\times$ 32 size of 10 categories, where the number of images in training and test set is 50,000 and 10,000, respectively. Similar to CIFAR-10, the number of categories in CIFAR-100 is 100. ImageNet contains 1,431,167 images with 224 $\times$ 224 size of 1,000 categories, where the number of images in training, validation and test set is 1,281,167, 50,000 and 100,000, respectively. Besides, Imagenette2 and Imagewoof2 are subsets of 10 easily classified classes from ImageNet. On CIFAR-10 and CIFAR-100, we train the networks on the training sets and evaluate the performance of the networks on the test sets. On ImageNet, Imagenette2 and Imagewoof2, we train on the training set and evaluate on the validation set. Besides, we adopt the accuracy to evaluate the performance and compare the FLOPs and parameters reductions to assess the acceleration and compression capability of pruned networks on all the datasets. For fair comparison, we reproduce the compared methods to demonstrate their performance. 

In addition, we apply our layer pruning method on large language models (LLMs), on the purpose of demonstrating the effectiveness of \textit{SGLP} on complex tasks and datasets. Specifically, we employ LLaMA3.1-18B-It as the backbone and conduct extensive layer pruning experiments with accuracy as the evaluation metric on \textbf{Reasoning:} PIQA and HellaSwag, \textbf{Examination:} ARC-easy (ARC-e), ARC-challenge (ARC-c), MMLU and CMMLU, \textbf{Resolution:} WinoGrande and \textbf{Comprehension:} OpenbookQA. Since LLM is commonly used for multi-tasking, we evaluate the importance of layers partitioned by Fisher Optimal Segmentation with perplexity (PPL), instead of GradNorm introduced in Sec. \ref{sec:gradnorm}.

\subsubsection{Configurations}

For image classification, we set a base learning rate of $0.01$ with SGD optimizer. Besides, the batch size is 256 and weight decay is $5 \times 10^{-4}$. In addition, we adopt DALI to accelerate data pre-processing. In terms of the experiments for LLMs, we set the rank as $8$ for LoRA to finetune the pruned models. Besides, we set the learning rate as $10^{-4}$ with warming steps as $100$ and leverage AdamW as the optimizer. We set the training batch size from $\{64, 128\}$ and the epoch as $2$. The experiments are conducted under PyTorch and with four NVIDIA A100 GPUs (40GB).

\subsubsection{Implementation Details}

In terms of Fisher Optimal Segmentation, several key parameters need to be carefully set to ensure the optimal partitioning of the data. Here, we outline the essential parameters and their settings:

The number of segments, denoted as $k$, is a critical hyperparameter that significantly influences the outcome of the segmentation. It determines how many distinct groups the data will be divided into. The choice of $k$ depends on the specific application and the nature of the data. In our implementation, $k$ is selected based on the depth of the neural network model:
\begin{itemize}
  \item For shallower networks (e.g., ResNet-20 or ResNet-32), a smaller $k$ is chosen to maintain a higher granularity of layer representation.
  \item For deeper networks (e.g., ResNet-110), a larger $k$ is used to balance the number of layers per segment and prevent excessively large segments that could lead to an explosion in the combinatorial search space during pruning.
\end{itemize}
As a general guideline, we suggest setting $k=3$ for most scenarios unless the network depth necessitates adjustments.

\subsection{Results and Analysis}

\subsubsection{Results on CIFAR-10 and CIFAR-100}

Tab. \ref{tab:cifar10-100} demonstrates the performance for our proposed \textit{SGLP} and other state-of-the-art methods in layer pruning on CIFAR-10 and CIFAR-100. The listed methods are arranged in descending order of FLOPs reduction ratios.

\begin{table}[t]
	\caption{Comparison of layer pruning methods on CIFAR-10 and CIFAR-100. The other tables follow the same convention.}
	\scalebox{1.0}{
		\centering 
		\begin{tabular}{lccc}
			\toprule
			\multicolumn{1}{l|}{Method}  & Acc. (\%) &  FLOPs $\downarrow (\%)$  & Params. $\downarrow (\%)$ \\
			\midrule
			
			\multicolumn{4}{c}{CIFAR-10 - VGG-16} \\
			\midrule
			
			\multicolumn{1}{l|}{Chen \etal~\cite{representations}}  & 93.47 & 38.9 & 87.9 \\
			\multicolumn{1}{l|}{\textbf{Ours (SGLP)}} & \textbf{94.43} & 42.0 & 54.2 \\
			\midrule
			\multicolumn{1}{l|}{Wu \etal~\cite{efficient-layer-pruning}} & 93.25 & 52.5 & -- \\
			\multicolumn{1}{l|}{Hossain \etal~\cite{hossain2024novel}} & 92.74 & 54.8 & 36.2 \\
            \multicolumn{1}{l|}{\textbf{Ours (SGLP)}} & \textbf{93.48} & 60.1 & 39.9 \\

            \midrule
			
			\multicolumn{4}{c}{CIFAR-10 - ResNet-56} \\
			\midrule
			
			\multicolumn{1}{l|}{Chen \etal~\cite{representations}}  & 93.29 & 34.8 & 42.3 \\
			\multicolumn{1}{l|}{SR-init~\cite{sr-init}}  & 93.83 & 37.5 & 64.0 \\
			\multicolumn{1}{l|}{GAL~\cite{gal}}  & 93.38 & 38.0 & 12.0 \\
                \multicolumn{1}{l|}{Hossain \etal~\cite{hossain2024novel}} & 92.88 & 48.4 & 52.1 \\
			\multicolumn{1}{l|}{\textbf{Ours (SGLP)}} & \textbf{93.83} & 52.6 & 56.4 \\
			\multicolumn{1}{l|}{Zhang \etal~\cite{shallower}}  & 93.40 & 52.7 & 44.7 \\
                \multicolumn{1}{l|}{Wu \etal~\cite{efficient-layer-pruning}}  & 93.08 & 63.8 & -- \\
			\multicolumn{1}{l|}{Soltani \etal~\cite{information}}  & 91.30 & 66.6 & 73.0 \\
			
			\midrule
			
			\multicolumn{4}{c}{CIFAR-100 - ResNet-56} \\
			\midrule
			\multicolumn{1}{l|}{SR-init~\cite{sr-init}} & 71.00 & 26.3 & 37.7 \\
			\multicolumn{1}{l|}{Lu \etal~\cite{modularity}}  & \textbf{71.40} & 30.2 & 9.2\\
			\multicolumn{1}{l|}{\textbf{Ours (SGLP)}}  & 71.07 & 33.9 & 21.1 \\
			\multicolumn{1}{l|}{Chen \etal~\cite{representations}} & 69.78 & 38.3 & 36.1 \\
			\midrule

                \multicolumn{1}{l|}{Hossain \etal~\cite{hossain2024novel}} & 67.22 & 44.6 & 40.2 \\
			\multicolumn{1}{l|}{\textbf{Ours (SGLP)}}  & \textbf{70.03} & 48.9 & 41.0 \\
			
			\bottomrule
		\end{tabular}
	}
	\label{tab:cifar10-100}
\end{table}

Compared with the method \cite{representations} proposed by Chen et. al, our presented \textit{SGLP} reduces larger FLOPs (42.0\% \vs 38.9\%) but achieves higher accuracy (94.43\% \vs 93.47\%), which demonstrates that pruning layers via Centered Kernel Alignment combining with Fisher Optimal Segmentation boosts the performance of the compact network, providing a more reliable basis for investigating unimportant layers. While under FLOPs reduction around 55\%, our proposed method still performs better than the method \cite{efficient-layer-pruning} and \cite{hossain2024novel} in accuracy (93.48\% \vs 93.25\% by \cite{efficient-layer-pruning} and 92.74 by \cite{hossain2024novel}), while yields larger FLOPs drop (60.1\% \vs 52.5\% by \cite{efficient-layer-pruning} and 54.8\% by \cite{hossain2024novel}). Thus, our proposed \textit{SGLP} shows its superiority of compressing and accelerating a neural network with plain structure.

On CIFAR-10, ResNet-56 is a relatively popular neural network, thus more methods are listed for comparison. Compared with GAL \cite{gal} and the method \cite{hossain2024novel}, our \textit{SGLP} removes remarkably larger proportion of parameters (56.4\% \vs 12.0\% by GAL and 52.1\% by \cite{hossain2024novel}). Besides, our method also reduces greatly more FLOPs than the work \cite{representations} and SR-init \cite{sr-init} (52.6\% \vs 34.8\% by \cite{representations} and 37.5\% by SR-init). In contrast to the method \cite{shallower}, our proposed \textit{SGLP} yields close FLOPs reduction (52.6\% \vs 52.7\%) and larger parameters drop (56.4\% \vs 44.7\%). In addition, our method achieves slightly lower FLOPs reduction than the approach introduced by Wu \etal \cite{efficient-layer-pruning} (63.8\%) and Soltani \etal \cite{information} (66.6\%). Therefore, our proposed \textit{SGLP} can compress the neural networks in layer level not only with plain structure like VGGNet but also with residual module.

In recent years, there have been relatively few layer pruning methods for CIFAR-100, so only two SOTAs and our proposed method are listed to demonstrate the performance for compressing ResNet-56. When achieving similar accuracy by SR-init \cite{sr-init} (71.00\%), the method \cite{modularity} (71.40\%) and our proposed \textit{SGLP} (71.07\%), our method reduces the largest FLOPs (33.9\% \vs 26.3\% by SR-init and 30.2\% by \cite{modularity}). However, the method \cite{representations} achieves relatively lower accuracy of 69.78\% with FLOPs and parameters reductions of 38.3\% and 36.1\%, respectively. Besides, under larger FLOPs and parameters drop of 48.9\% and 41.0\%, respectively, our method still yields competitive Top-1 accuracy of 70.03\%, larger than the method \cite{hossain2024novel} of 67.22\%. Therefore, our method still achieves excellent performance on more complex datasets, which demonstrates the competitiveness of our method by evaluating the importance of layers based on the gradients during back-propagation. 

\subsubsection{Results on ImageNet}

Tab. \ref{tab:imagenet} demonstrates the performance for our proposed \textit{SGLP} and other state-of-the-art methods in layer pruning for ResNet-50 on ImageNet. Generally, our \textit{SGLP} outperforms its counterparts in accuracy. Besides, we observe that the method \cite{elkerdawy2020one} achieves the Top-1 accuracy of 74.74\%, while it only removes 8.24\% of parameters. Furthermore, SR-init \cite{sr-init} achieves similar accuracy (75.41\%), but shows advantages over FLOPs reduction (15.59\%) and parameters reduction (39.29\%). Nevertheless, our proposed \textit{SGLP} advances all the methods in accuracy of 75.71\%, and achieves FLOPs and parameters reductions of 21.30\% and 30.60\%, respectively. Hence, experiments on ImageNet demonstrate that the effectiveness of our proposed \textit{SGLP} lies in its ability to find the optimal layer combinations for pruning by combining CKA, Fisher Optimal Segmentation, and GradNorm. This combination not only reduces the size and computational footprint of the network but also ensures that the pruned network maintains high accuracy, making it an effective solution for deploying deep neural networks on resource-limited platforms on large and complex datasets.

\begin{table}[t]
	\caption{Comparison of layer pruning methods for ResNet-50 on ImageNet.}
	\scalebox{1.0}{
		\centering 
		\begin{tabular}{lccc}
			\toprule
			\multicolumn{1}{l|}{Method}  & Acc. (\%) &  FLOPs $\downarrow (\%)$  & Params. $\downarrow (\%)$ \\
			\midrule
			
			\multicolumn{1}{l|}{Elkerdawy \etal~\cite{elkerdawy2020one}}  & 74.74 & -- & 8.24 \\
			\multicolumn{1}{l|}{SR-init~\cite{sr-init}}  & 75.41 & 15.59 & 39.29 \\
			\multicolumn{1}{l|}{\textbf{Ours (SGLP)}} & \textbf{75.71} & 21.30 & 30.60 \\
			
			\bottomrule
		\end{tabular}
	}
	\label{tab:imagenet}
\end{table}

\subsubsection{Results on Imagenette2 and Imagewoof2}
Tab. \ref{tab:imagenette2 and imagewoof2} demonstrates the performance for our proposed \textit{SGLP} and other state-of-the-art methods in layer pruning on ImageNet branch datasets, Imagenette2 and Imagewoof2. 

On ImageNette2, in general, all the listed methods yield similar FLOPs and parameters reductions. Among them, SR-init \cite{sr-init} achieves accuracy lower than 90.00\%. In contrast, the method \cite{elkerdawy2020one} and our proposed \textit{SGLP} maintain the accuracy over 90.00\%, while our method achieves better performance (90.42\% \vs 90.12\%). 

Similar to the cases on ImageNette2, the methods compared also achieve close FLOPs and parameters drop on Imagewoof2. Nevertheless, the methods except our \textit{SGLP} all yield the accuracy under 85.00\% (84.13\% by \cite{elkerdawy2020one} \vs 84.73\% by SR-init). On the contrary, our proposed \textit{SGLP} outperforms other methods in accuracy of 85.16\%, while reducing 21.33\% FLOPs and removing 15.43\% parameters.

Experiments on these two datasets show that our proposed \textit{SGLP} has great potential in compressing neural networks in image classification. Therefore, it demonstrates the superiority of our approach to investigate the internal importance of the layers in a network by Centered Kernel Alignment (CKA) and remove the unimportant ones with Fisher Optimal Segmentation and GradNorm.

\begin{table}[t]
	\caption{Comparison of layer pruning methods on Imagenette2 and Imagewoof2 for ResNet-50.}
	\scalebox{1.0}{
		\centering 
		\begin{tabular}{lccc}
			\toprule
			\multicolumn{1}{l|}{Method}  & Acc. (\%) &  FLOPs $\downarrow (\%)$  & Params. $\downarrow (\%)$ \\
			\midrule
			
			\multicolumn{4}{c}{Imagenette2} \\
			\midrule
			
			\multicolumn{1}{l|}{Elkerdawy \etal~\cite{elkerdawy2020one}}  & 90.12 & 17.32 & 7.98 \\
			\multicolumn{1}{l|}{SR-init~\cite{sr-init}}  & 89.34 & 20.25 & 6.91 \\
			\multicolumn{1}{l|}{\textbf{Ours (SGLP)}} & \textbf{90.42} & 21.34 & 8.81 \\
			\midrule  
			\multicolumn{4}{c}{Imagewoof2} \\
			\midrule	
			\multicolumn{1}{l|}{Elkerdawy \etal~\cite{elkerdawy2020one}}  & 84.13 & 19.29 & 12.68 \\
			\multicolumn{1}{l|}{SR-init~\cite{sr-init}}  & 84.73 & 18.73 & 13.75 \\
			\multicolumn{1}{l|}{\textbf{Ours (SGLP)}} & \textbf{85.16} & 21.33 & 15.43 \\
			\bottomrule
		\end{tabular}
	}
	\label{tab:imagenette2 and imagewoof2}
\end{table}

Our method's effectiveness stems from the strategic integration of Centered Kernel Alignment (CKA) and Fisher Optimal Segmentation within the RSNP framework. CKA allows us to quantify the representational similarities between layers, providing a data-driven basis for layer analysis and selection. This is crucial as it aids in identifying and removing redundant layers that exhibit high overlap in their feature representations without compromising the network's performance.

By applying Fisher Optimal Segmentation based on the CKA-derived similarity matrix, we partition the network into segments with high intra-segment similarity and distinct inter-segment differences. This approach effectively reduces the search space for pruning, focusing on the layers within these segments rather than conducting importance evaluation across all layers of the network, thereby reducing the computational cost and improving the efficiency of the layer pruning process.

The changes in network structure after pruning, such as the reduction in layer depth and parameter count, are carefully analyzed for their impact on the model's generalization. We observe that pruning can act as a regularizer, preventing overfitting by focusing on essential features and reducing redundancy. This leads to a model that not only maintains accuracy but also generalizes better to unseen data. Our experiments show that certain models can be pruned to high sparsity levels while outperforming their dense counterparts, indicating that pruning can enhance model robustness and efficiency.

\subsubsection{Results for Large Language Models}

The quantitative results for LLaMA3.1-8B-It on multiple benchmark datasets are listed in Tab. \ref{tab:llm}, where the compared algorithms are reproduced by us. In general, our proposed \textit{SGLP} outperforms its counterparts in accuracy on all the datasets. All the experiments are conducted under a pruning ratio of 25\%. 

From the results, we observe that the Random and Magnitude-based pruning methods (L1-norm and L2-norm) lead to significant performance degradation across all benchmarks. For instance, on PIQA, random pruning only achieves 56.53\% accuracy compared to 80.03\% (unpruned). Similarly, on more challenging tasks like OpenbookQA and ARC-c, random pruning results in sharp drops to 14.00\% and 18.60\%, respectively. The performance of L1-norm and L2-norm magnitude-based methods is also quite poor, with average accuracies of 30.18\% and 30.15\%, respectively, indicating that these approaches indiscriminately remove important layers, leading to reduced model capacity.

In contrast, ShortGPT shows considerable improvements over the random and magnitude-based approaches, particularly on tasks like PIQA (71.76\%) and HellaSwag (41.96\%). However, it still lags behind the Dense model significantly, especially on more knowledge-intensive benchmarks like MMLU and CMMLU, where it only achieves 24.17\% and 24.94\%, respectively.

Our proposed method, \textit{SGLP}, outperforms all the pruning methods across every benchmark, demonstrating superior robustness and balance between pruning efficiency and model performance. For example, \textit{SGLP} achieves 74.59\% on PIQA, 47.15\% on HellaSwag, and 36.95\% on ARC-c, surpassing ShortGPT by 2-4 percentage points across most tasks. Moreover, \textit{SGLP} performs particularly well on the MMLU and CMMLU benchmarks, where it reaches 35.22\% and 28.98\%, respectively, significantly improving over other methods. Overall, \textit{SGLP} achieves the highest average accuracy of 46.86\%, closely approximating the unpruned model's 62.99\%. Therefore, the excellent results demonstrate the effectiveness of our layer pruning method when applied not only in image classification but also on large language models. Hence, our proposed \textit{SGLP} has great potential for deploying large language models in resource-constrained environments without a substantial performance loss.

\begin{table*}[t]
	\caption{Comparison of layer pruning methods for LLaMA3.1-8B-It across various benchmarks. ``Dense'' refers to the unpruned LLM, and ``L1'' and ``L2'' represent L1-norm and L2-norm, respectively.}
	\scalebox{1.17}{
		\centering 
		\begin{tabular}{l|cccccccc|c}
			\toprule
			\multirow{2}{*}{Method} & \multicolumn{8}{c|}{Benchmark} & \multirow{2}{*}{Avg Acc.} \\
			\cmidrule{2-9}
			& PIQA & HellaSwag & OpenbookQA & ARC-e & ARC-c & MMLU & CMMLU & WinoGrande &  \\
			\midrule
			
			\multicolumn{1}{l|}{Dense}  & 80.03 & 59.10 & 33.80 & 81.82 & 51.79 & 67.90 & 55.52 & 73.95 & 62.99 \\
			
			\midrule
			
			\multicolumn{1}{l|}{Random}  & 56.53 & 28.86 & 14.00 & 31.69 & 18.60 & 22.75 & 25.59 & 50.75 & 31.10 \\
			\multicolumn{1}{l|}{Magnitude-L1} & 54.08 & 26.34 & 13.60 & 28.45 & 20.14 & 25.04 & 25.03 & 48.78 & 30.18 \\
			\multicolumn{1}{l|}{Magnitude-L2} & 54.13 & 26.38 & 13.40 & 28.41 & 20.14 & 24.98 & 25.04 & 48.70 & 30.15 \\
			\multicolumn{1}{l|}{ShortGPT} & 71.76 & 41.96 & 20.20 & 61.07 & 28.41 & 24.17 & 24.94 & 53.91 & 40.80 \\
			\multicolumn{1}{l|}{\textbf{Ours (SGLP)}} & \textbf{74.59} & \textbf{47.15} & \textbf{27.00} & \textbf{63.93} & \textbf{36.95} & \textbf{35.22} & \textbf{28.98} & \textbf{61.09} & \textbf{46.86} \\
			
			\bottomrule
		\end{tabular}
	}
	\label{tab:llm}
\end{table*}

\subsubsection{Results for Signal Classification Models}
 
Tab. \ref{tab:results_signal} provides a comparative analysis of different layer pruning methods applied to CNN1D models across three datasets: RML2016.10a \cite{o2016radio}, Sig2019-12 \cite{chen2021signet}, and RML2018.01a \cite{o2018over}, all at a pruning rate of 67\%. The performance of each pruning method is assessed based on the preservation of accuracy post-pruning, as well as the reduction in computational resources, specifically measured by the decrease in FLOPs and parameters. The table indicates that while all methods experience a drop in accuracy due to pruning, the proposed \textit{SGLP} manages to maintain or even improve upon the accuracy of other pruning techniques in some cases, such as achieving a slightly higher accuracy on RML2016.10a and Sig2019-12 datasets compared to random pruning and other methods like LCP and SR-init. Additionally, SGLP demonstrates a consistent reduction in FLOPs and parameters, which aligns with the goal of pruning to optimize models for deployment on resource-constrained devices. The results suggest that SGLP is effective in removing redundant layers without significantly compromising the model's predictive power, highlighting its potential and robustness for practical applications where model efficiency is critical.

\begin{table*}[t]
{
    \centering
    \caption{Comparison of layer pruning methods for RML2016.10a, Sig2019-12 and RML2018.01a with CNN1D.}
\resizebox{\linewidth}{!}{
    \begin{tabular}{cccc|cc|cc}\toprule
Model&Dataset&Pruning Rate&Method&Original Acc (\%)&Acc (\%)&FLOPs PR&Params PR\\ \midrule
\multirow{12}{*}{CNN1D}&\multirow{4}{*}{RML2016.10a}&\multirow{4}{*}{67\%}&Random&\multirow{4}{*}{59.45}&57.67 &8.3\%&52.8\%\\
&&&LCP \cite{lcp}&&58.85 &8.3\%&52.8\%\\
&&&SR-init \cite{sr-init}&&58.00 &8.3\%&52.8\%\\
&&&\textbf{Ours (SGLP)}&&\bf  59.16 &8.3\%&52.8\%\\ \cmidrule{2-8}
&\multirow{4}{*}{Sig2019-12}&\multirow{4}{*}{67\%}&Random&\multirow{4}{*}{64.51} &59.88 &8.4\%&42.4\%\\
&&&LCP \cite{lcp}&&59.81 &8.4\%&42.4\%\\
&&&SR-init \cite{sr-init} &&59.91 &8.4\%&42.4\%\\
&&&\textbf{Ours (SGLP)}&&\bf 60.39 &8.4\%&42.4\%\\ \cmidrule{2-8}
&\multirow{4}{*}{RML2018.01a}&\multirow{4}{*}{67\%}&Random&\multirow{4}{*}{84.16}&75.89 &8.4\%&33.3\%\\
&&&LCP \cite{lcp}&&80.93 &8.4\%&33.3\%\\
&&&SR-init \cite{sr-init}&&81.90 &8.4\%&33.3\%\\
&&&\textbf{Ours (SGLP)}&&\bf 82.65 &8.4\%&33.3\% \\ \bottomrule
    \end{tabular}}
    \label{tab:results_signal}
}
\end{table*}

\subsection{Ablation Studies}

\subsubsection{Impact of Proposed Components}

Tab. \ref{tab:ablation_complexity} offers a detailed analysis of the search space, evaluation cost, and performance when applying different components of our presented \textit{SGLP} for pruning ResNet-56 on CIFAR-10. The table compares the impact of using Centered Kernel Alignment (CKA) as the layer similarity, the employment of Fisher Optimal Segmentation for network partitioning, and the utilization of GradNorm for layer importance evaluation on the accuracy of the pruned network.

The results indicate that the integration of all three components—CKA, segmentation, and GradNorm—into the SGLP framework leads to the highest accuracy of 93.83\%, demonstrating the effectiveness of the comprehensive approach. When segmentation is omitted, the accuracy drops to 93.05\%, highlighting the importance of network partitioning in identifying redundant layers. Similarly, excluding GradNorm results in a slightly lower accuracy of 93.08\%, underscoring the value of efficient layer importance assessment.

In terms of the search space, the table reveals that using CKA and segmentation significantly reduces the number of possible combinations, which is a substantial decrease. This reduction is crucial for managing the computational complexity of the pruning process. Furthermore, the evaluation cost is markedly lower when all components are used together, taking approximately 1 hour compared to over 3000 hours without segmentation, illustrating the significant time savings afforded by our proposed \textit{SGLP}.

\begin{table}[htbp]
    \centering
{
    \caption{Searching space and evaluation cost, as well as performance with and without different proposed modules for pruning ResNet-56 on CIFAR-10. ``inf'' denotes infinite costs.}
\scalebox{0.85}{
    \begin{tabular}{ccc|c|cc}\toprule
CKA&Segmentation&GradNorm&Acc.&Search Space&Evaluation Cost\\ \midrule
\checkmark &&&93.81&$2^{28}-1$&inf\\
&\checkmark &&93.05&$ 16430 $&3000+ h\\
&&\checkmark &93.08&$2^{28}-1$&inf\\
\checkmark &\checkmark &\checkmark &93.83& 16430 &$\sim$1h\\ \bottomrule
    \end{tabular}}
    \label{tab:ablation_complexity}
}
\end{table}
\subsubsection{Impact of k}

\begin{figure}[htbp]
	\centering
	\subfigure[Performance of the pruned networks under different $k$ for pruning ResNet-56 on CIFAR-10.]{
		\begin{minipage}[t]{0.48\textwidth}
			\centering
			\includegraphics[scale=0.48]{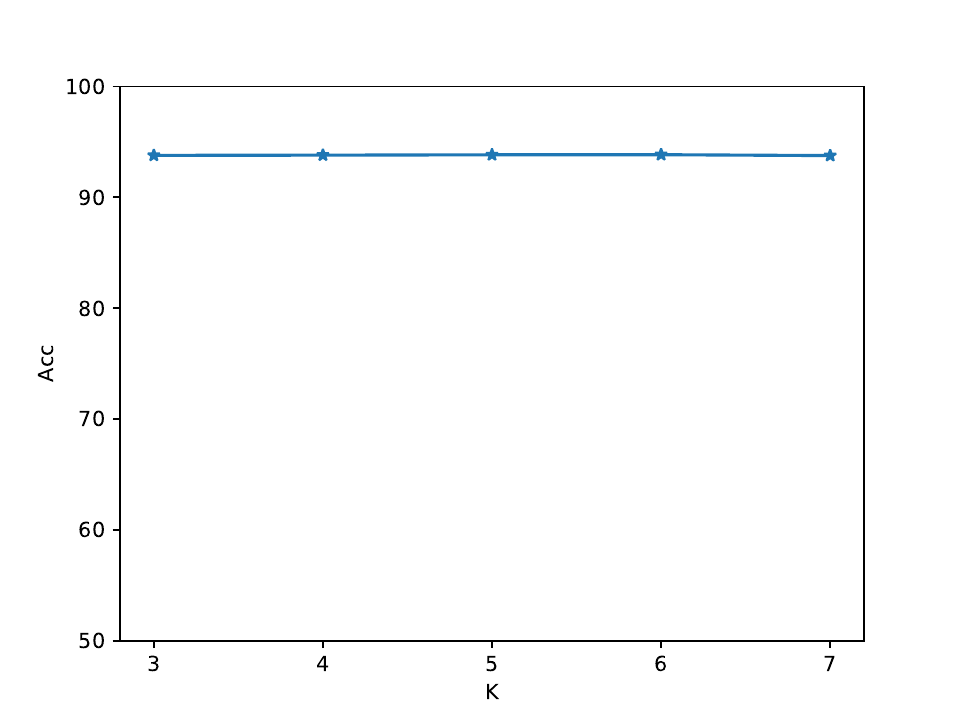}
			\label{fig:k-acc}
		\end{minipage}
	}
	\subfigure[Performance of the pruned networks when preserving different layers.]{
		\begin{minipage}[t]{0.48\textwidth}
			\centering
			\includegraphics[scale=0.48]{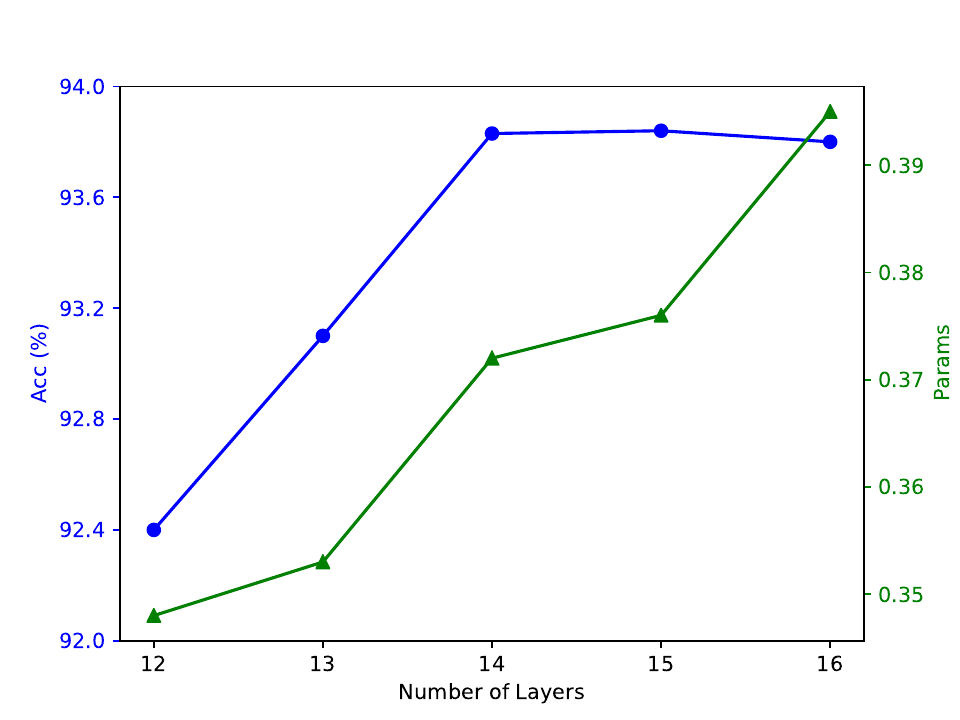}
			\label{fig:layer-deletion}
		\end{minipage}
	}
	\centering
	\caption{Results of ablation studies.}
\end{figure}

Sec. \ref{sec:fisher} mentions that $k$ is an important hyper-parameter that needs heuristic tuning for the layer segmentation. A smaller \(k\) in deep models results in larger segments, which may lead to an explosion in the combinatorial search space during pruning. Hence, increasing \(k\) appropriately helps keep the search space manageable. Thus, by tuning \(k\) based on the depth of the model, we ensure consistent performance without imposing excessive computational burdens. We investigate the performance of the pruned ResNet-56 on CIFAR-10 which are partitiond into different $k$ segments while maintaining the same number of layers. Fig. \ref{fig:k-acc} demonstrates that under different $k$, pruned networks with consistent number of remained layers achieve rather close performance. The results indicate a consistent level of accuracy across different $k$ values, suggesting that the Similarity Guided Fast Layer Partition Pruning (SGLP) method can reliably identify redundant layers regardless of how the network is segmented. This consistency implies that the pruning process is robust and does not overly depend on the specific number of segments, offering flexibility in choosing k based on computational constraints or architectural considerations. The flat trend in accuracy also points to the generalization capability of the pruned models, as they maintain performance across different partitionings, indicating that the pruning strategy is not overfitting to a particular structure but is preserving the critical features necessary for classification tasks. As a result, we suggest setting \(k\)=3 for most scenarios unless the network depth necessitates adjustments.

Fig. \ref{fig:layer-deletion} investigates the relationship between the number of layers preserved in the pruned network and both the Top-1 accuracy and the parameter count. The analysis reveals an initial increase in accuracy as more layers are retained, which is expected since additional layers can capture more intricate features beneficial for classification. However, this improvement plateaus after a certain point, around 15 layers as observed in the experiment, suggesting that beyond this threshold, the added layers do not significantly enhance the network's performance. This could be attributed to the network reaching its capacity for learning from the training data or even beginning to overfit. Concurrently, the parameter count increases with the number of layers, but the rate of increase in accuracy slows down, indicating a point of diminishing returns. This suggests that there is an optimal number of layers beyond which the network does not benefit significantly in terms of performance, highlighting the importance of finding a balance between model complexity and efficiency. Our \textit{SGLP} aids in identifying this balance by pruning away layers that do not contribute substantially to performance while maintaining accuracy, thus regularizing the network and potentially reducing overfitting.

\subsubsection{Impact of Batch Size}

To address the potential impact of batch size on the performance of our layer pruning method, we conducted a set of experiments to systematically evaluate how different batch sizes affect the accuracy, computational efficiency, and parameter reduction of our model.

As shown in Tab. \ref{tab:batch_size_performance}, the accuracy of the model remains relatively stable across different batch sizes, with only minor fluctuations observed. This indicates that our layer pruning method is not sensitive to the choice of batch size, which is a desirable property for ensuring consistent performance in various deployment scenarios. The FLOPs reduction and parameter reduction also show only slight variations, suggesting that the efficiency gains from our pruning approach are maintained regardless of the batch size used during training.

\begin{table}[htbp]
    \centering
    \caption{Performance for pruning ResNet-56 CIFAR-10 with different batch sizes.}
    \begin{tabular}{ccccc}
        \toprule
        Batch Size & Acc. (\%) & FLOPs $\downarrow$ (\%) & Params. $\downarrow$ (\%) \\
        \midrule
        128 & 93.83 & 52.6 & 56.4 \\
        256 & 93.81 & 53.2 & 58.5 \\
        512 & 93.79 & 54.6 & 56.8 \\
        1024 & 93.88 & 52.9 & 55.4 \\
        \bottomrule
    \end{tabular}
    \label{tab:batch_size_performance}
\end{table}

\subsubsection{Performance w.r.t Number of Layers}

For pruning ResNet-56 on CIFAR-10, the pruned model listed in Tab. \ref{tab:cifar10-100} remains 14 layers finally. Fig. \ref{fig:layer-deletion} shows the performance and parameters of the pruned network when remaining different number of layers for pruning ResNet-56 on CIFAR-10 with 5 segments, where the x-axis represents the number of layers, while the left y-axis indicates the Top-1 accuracy and the right y-axis denotes the parameters of the networks. Initially, there is a noticeable improvement in accuracy as more layers are preserved, indicating that additional layers contribute positively to the network performance. It is evident that as the number of layers increases, the parameters of the pruned network also increase. This is expected since more layers would typically mean more parameters. However, the accuracy of the network does not increase proportionally to the increase in the number of layers. 

In addition, after the number of layer reaches 15, the improvement in accuracy starts to diminish and may even decline. This suggests that beyond the optimal number of layers (15 in our experiment), the additional layers do not provide significant benefits to the network performance, even the performance of the network starts to degrade. This could be due to over-fitting, where the model becomes too complex and starts to memorize the training data rather than learning to generalize from it. Thus, it highlights the importance of finding the right balance between the number of layers and the network performance, and demonstrates our \textit{SGLP} can assist us to find the best number of layers to remove redundant layers that do not contribute significantly to the performance while maintaining the network accuracy. Besides, our proposed method can provide a regularization effect for the network and reduce its over-fitting.

\subsubsection{Application of Our Method to Filter/Channel Pruning}

On the purpose of validating the effectiveness and universality of our algorithm, we transfer and apply our method into filter and channel pruning, where what is different from the existing pipeline is that we partition the channels into segments via the similarity between each two channel. Tab. \ref{tab:application} demonstrates the performance of our method \vs other methods in filter/channel pruning for VGG-16 and ResNet-56 on CIFAR-10.

For VGG-16, AFIE \cite{lu2024entropy} achieves an accuracy of 93.35\% with a 54.5\% reduction in FLOPs and a 43.1\% reduction in parameters. In contrast, our proposed method outperforms it in all metrics, achieving an accuracy of 93.96\%, with a 55.2\% reduction in FLOPs and a 46.8\% reduction in parameters. Thus, it demonstrates that our method not only achieves a higher accuracy but also reduces the computational and memory overhead more effectively in channel pruning.

When pruning ResNet-56, HRank \cite{hrank} reduces 50.0\% FLOPs and removes 42.4\% parameters, yielding Top-1 accuracy of 93.17\%. GAL \cite{gal} under sparsity of 0.6 achieves Top-1 accuracy of 92.98\% with FLOPs and parameters reductions of 37.6\% and 11.8\%, respectively. CHIP \cite{chip} achieves Top-1 accuracy of 94.16\% under FLOPs reduction of 47.4\% and parameters reduction of 42.8\%. Recent research RLAL \cite{ganjdanesh2024jointly} achieves an accuracy of 93.86\% with a 50.0\% reduction in FLOPs. In addition, UDSP \cite{gao2024bilevelpruning} yields an accuracy of 93.78\% with FLOPs and parameters reductions of 50.1\% and 20.0\%, respectively. On the contrary, our proposed \textit{SGLP} obtains Top-1 accuracy of 94.21\%, outperforming other compared methods with FLOPs and parameters reductions of 50.0\% and 42.1\%, respectively. 

Experiments in channel pruning demonstrate that the effectiveness and universality of our proposed \textit{SGLP}, thus our proposed method has great potential in compressing deep neural networks.

\begin{table}[h!]
\centering
	\caption{Performance of our method \vs other channel pruning methods for VGG-16 and ResNet-56 on CIFAR-10.}
	\scalebox{0.99}{
		\centering 
		\begin{tabular}{lccc}
			\toprule
			\multicolumn{1}{l|}{Method}  & Acc. (\%) &  FLOPs $\downarrow (\%)$  & Params. $\downarrow (\%)$ \\
			\midrule

                \multicolumn{4}{c}{CIFAR-10 - VGG-16} \\
			\midrule

                \multicolumn{1}{l|}{AFIE~\cite{lu2024entropy}}  & 93.35 & 54.5 & 43.1 \\
                \multicolumn{1}{l|}{\textbf{Ours (SGLP)}} & \textbf{93.96} & 55.2 & 46.8 \\

                \midrule
                
                \multicolumn{4}{c}{CIFAR-10 - ResNet-56} \\
			\midrule
            
			\multicolumn{1}{l|}{HRank~\cite{hrank}}  & 93.17 & 50.0 & 42.4 \\
			\multicolumn{1}{l|}{GAL-0.6~\cite{gal}}  & 92.98 & 37.6 & 11.8 \\
			\multicolumn{1}{l|}{CHIP~\cite{chip}}  & 94.16 & 47.4 & 42.8 \\
                \multicolumn{1}{l|}{RLAL~\cite{ganjdanesh2024jointly}}  & 93.86 & 50.0 & -- \\
                \multicolumn{1}{l|}{UDSP~\cite{gao2024bilevelpruning}}  & 93.78 & 50.1 & 20.0 \\
            \multicolumn{1}{l|}{\textbf{Ours (SGLP)}} & \textbf{94.21} & 50.0 & 42.1 \\
			
			\bottomrule
		\end{tabular}
	}
	\label{tab:application}
\end{table}

\subsubsection{Pruned and Remained Layers}

\begin{figure}[ht!]
	\centering
	\includegraphics[height=5cm,width=0.7\textwidth,keepaspectratio]{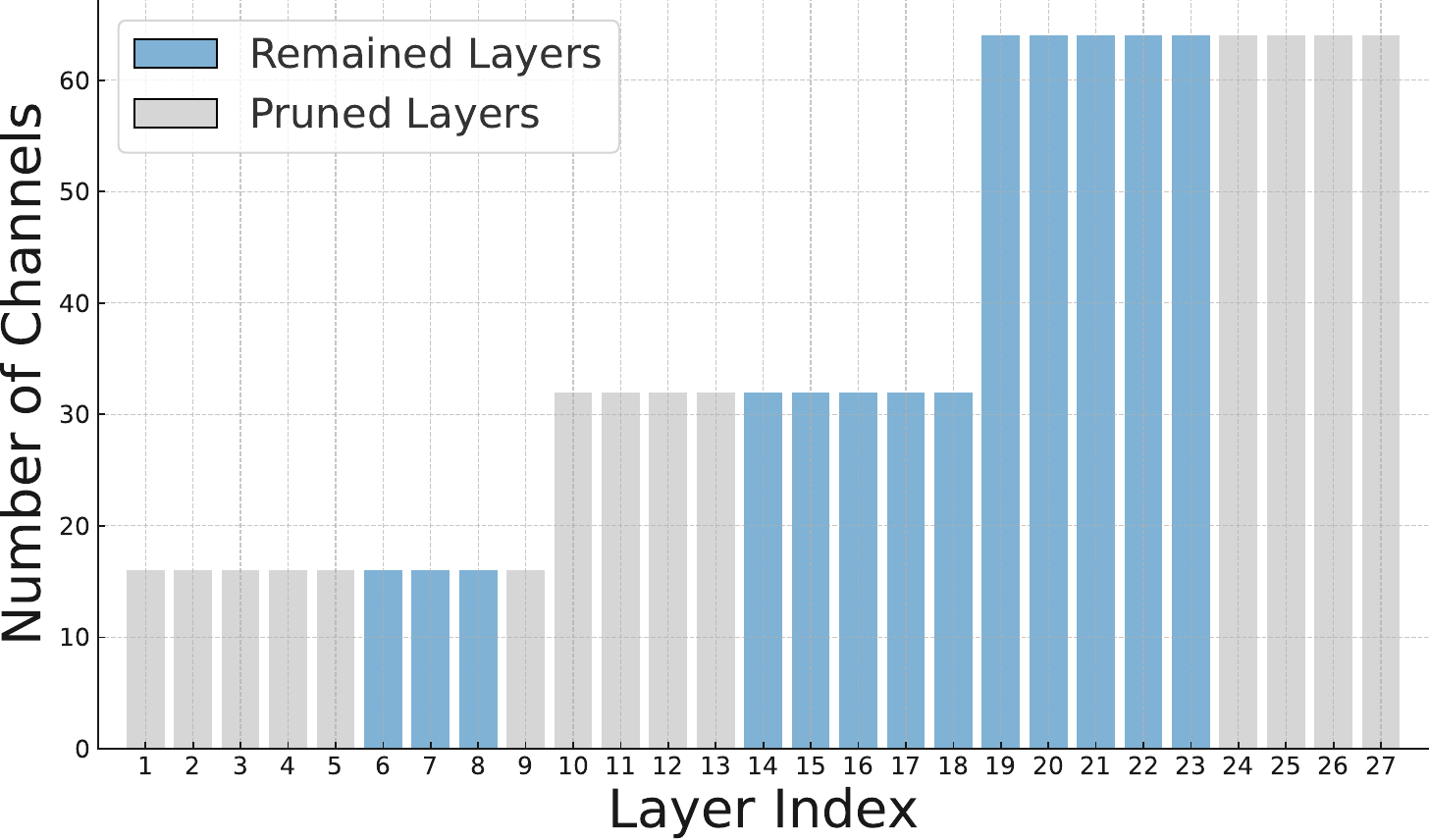}
	\caption{Pruned and remained layers for ResNet-56 on CIFAR-10.}
	\label{fig:pruned-layers}
\end{figure}

Fig. \ref{fig:pruned-layers} demonstrates the pruned and remained layers for ResNet-56 on CIFAR-10, where the x-axis represents the index of the layers, while the y-axis indicates the number of channels in each layer. The blue bars correspond to the layers that were kept, and the gray bars represent the layers that were pruned. In this experiment, each residual block is considered as an individual layer, allowing for a more granular analysis of the pruning effect. From the figure, it can be observed that our proposed \textit{SGLP} tends to prune channels from the shallower layers of the network, suggesting that these layers may be redundant and contribute less to the overall performance of the task. In contrast, deeper layers are more likely to be kept, indicating their critical role in feature extraction and representation.

\section{Discussion and Conclusion}

In this study, we proposed Similarity-Guided fast Layer Partition pruning (SGLP), a novel layer pruning framework designed to reduce the computational complexity of Deep Neural Networks (DNNs) while preserving their performance. By leveraging Centered Kernel Alignment (CKA) to measure representational similarities among layers and employing Fisher Optimal Segmentation to partition the network into semantically coherent segments, our SGLP enables more informed and structured pruning decisions. Within each segment, we further apply GradNorm to efficiently estimate layer importance without requiring fine-tuning, streamlining the pruning process and significantly improving practicality. Extensive experiments across diverse tasks including image classification and LLMs, demonstrate that our proposed method consistently outperforms existing state-of-the-art methods in both accuracy retention and computational efficiency. These results highlight the effectiveness of our approach in identifying redundant layers while preserving the functional integrity of the pre-trained network.

Our method offers several significant strengths. First, by incorporating representation similarity into the pruning process, SGLP ensures that critical information flow across layers is maintained, leading to more robust compressed models. Second, segment-wise pruning reduces the granularity of decisions, enabling more stable and targeted layer removal. Third, the use of GradNorm for importance evaluation avoids costly fine-tuning, making SGLP both faster and more deployable in real-world scenarios.

Nevertheless, our approach also has limitations. Its effectiveness relies on the quality of the pre-trained model, and determining the optimal number of segments in Fisher segmentation requires careful tuning. Furthermore, while SGLP is currently designed for sequential network structures, its application to more complex architectures such as multi-branch or attention-based models may necessitate additional adaptation.

Our work contributes a new perspective to the field of network compression by combining similarity analysis, optimal segmentation, and efficient layer evaluation into a unified framework. This approach provides a practical and scalable solution for deploying large-scale DNNs on resource-limited platforms, a critical need in edge AI and mobile computing. 

In future work, we aim to extend SGLP to support a broader range of neural architectures, including convolutional and transformer-based models with non-linear or multi-path structures. We also plan to explore automated hyperparameter tuning strategies and investigate how SGLP can be integrated with other compression techniques, such as quantization or knowledge distillation, to achieve even higher compression ratios. Moreover, we will validate the generalizability of SGLP on additional deep learning tasks beyond image classification and LLMs, such as object detection, semantic segmentation, and other vision or language applications. 

In summary, our proposed SGLP method represents a significant advancement in layer pruning for DNNs, providing a practical solution for deploying large models on devices with limited computational resources.

\bibliographystyle{ieeetr}
\bibliography{references}

@inproceedings{resnet,
  title={Deep residual learning for image recognition},
  author={He, Kaiming and Zhang, Xiangyu and Ren, Shaoqing and Sun, Jian},
  booktitle={Proceedings of the IEEE conference on computer vision and pattern recognition},
  pages={770--778},
  year={2016}
}

@article{ou2025social,
  title={Social Media as an Emotional Barometer: Bidirectional Encoder Representations From Transformers--Long Short-Term Memory Sentiment Analysis on the Evolution of Public Sentiments During Influenza A on Sina Weibo},
  author={Ou, Yifan and de Bruijn, Gert-Jan and Schulz, Peter Johannes},
  journal={Journal of Medical Internet Research},
  volume={27},
  pages={e68205},
  year={2025},
  publisher={JMIR Publications Toronto, Canada}
}

@inproceedings{ou2025analyzing,
  title={Analyzing user behavior in online communities using data crawling and machine learning algorithms (Erratum)},
  author={Ou, Yifan and Sun, Feifei and Tan, Cheng and Shi, Haoran and Zhang, Mengyang and Gong, Chen},
  booktitle={Fifth International Conference on Telecommunications, Optics, and Computer Science (TOCS 2024)},
  volume={13629},
  pages={785--785},
  year={2025},
  organization={SPIE}
}

@article{li2025sepprune,
  title={Sepprune: Structured pruning for efficient deep speech separation},
  author={Li, Yuqi and Li, Kai and Yin, Xin and Yang, Zhifei and Dong, Junhao and Dong, Zeyu and Yang, Chuanguang and Tian, Yingli and Lu, Yao},
  journal={arXiv preprint arXiv:2505.12079},
  year={2025}
}

@inproceedings{cka,
  title={Similarity of neural network representations revisited},
  author={Kornblith, Simon and Norouzi, Mohammad and Lee, Honglak and Hinton, Geoffrey},
  booktitle={International conference on machine learning},
  pages={3519--3529},
  year={2019},
  organization={PMLR}
}

@article{lcp,
  author={Lu, Yao and Zhu, Yutao and Li, Yuqi and Xu, Dongwei and Lin, Yun and Xuan, Qi and Yang, Xiaoniu},
  journal={IEEE Transactions on Cognitive Communications and Networking}, 
  title={A Generic Layer Pruning Method for Signal Modulation Recognition Deep Learning Models}, 
  year={2024},
}

@inproceedings{pfec,
  title={Pruning filters for efficient convnets},
  author={Li, Hao and Kadav, Asim and Durdanovic, Igor and Samet, Hanan and Graf, Hans Peter},
  booktitle={International Conference on Learning Representations},
  pages={1--12},
  year={2017}
}

@inproceedings{fpgm,
  title={Filter pruning via geometric median for deep convolutional neural networks acceleration},
  author={He, Yang and Liu, Ping and Wang, Ziwei and Hu, Zhilan and Yang, Yi},
  booktitle={Proceedings of the IEEE/CVF Conference on Computer Vision and Pattern Recognition},
  pages={4340--4349},
  year={2019}
}

@inproceedings{sr-init,
  title={SR-init: An Interpretable Layer Pruning Method},
  author={Tang, Hui and Lu, Yao and Xuan, Qi},
  booktitle={ICASSP 2023-2023 IEEE International Conference on Acoustics, Speech and Signal Processing (ICASSP)},
  pages={1--5},
  year={2023},
  organization={IEEE}
}

@article{efficient-layer-pruning,
  title={Efficient layer compression without pruning},
  author={Wu, Jie and Zhu, Dingshun and Fang, Leyuan and Deng, Yue and Zhong, Zhun},
  journal={IEEE Transactions on Image Processing},
  year={2023},
  publisher={IEEE}
}

@inproceedings{modularity,
  title={Understanding the dynamics of dnns using graph modularity},
  author={Lu, Yao and Yang, Wen and Zhang, Yunzhe and Chen, Zuohui and Chen, Jinyin and Xuan, Qi and Wang, Zhen and Yang, Xiaoniu},
  booktitle={European Conference on Computer Vision},
  pages={225--242},
  year={2022},
  organization={Springer}
}

@article{lu2024reassessing,
  title={Reassessing layer pruning in llms: New insights and methods},
  author={Lu, Yao and Cheng, Hao and Fang, Yujie and Wang, Zeyu and Wei, Jiaheng and Xu, Dongwei and Xuan, Qi and Yang, Xiaoniu and Zhu, Zhaowei},
  journal={arXiv preprint arXiv:2411.15558},
  year={2024}
}

@inproceedings{hrank,
  title={Hrank: Filter pruning using high-rank feature map},
  author={Lin, Mingbao and Ji, Rongrong and Wang, Yan and Zhang, Yichen and Zhang, Baochang and Tian, Yonghong and Shao, Ling},
  booktitle={Proceedings of the IEEE/CVF conference on computer vision and pattern recognition},
  pages={1529--1538},
  year={2020}
}

@inproceedings{cicc,
  title={CICC: Channel Pruning via the Concentration of Information and Contributions of Channels.},
  author={Chen, Yihao and Li, Zhishan and Yang, Yingqing and Xie, Lei and Liu, Yong and Ma, Longhua and Liu, Shanqi and Tian, Guanzhong},
  booktitle={BMVC},
  pages={243},
  year={2022}
}

@article{obd,
  title={Optimal brain damage},
  author={LeCun, Yann and Denker, John and Solla, Sara},
  journal={Advances in neural information processing systems},
  volume={2},
  year={1989}
}

@article{comparing,
  title={Comparing biases for minimal network construction with back-propagation},
  author={Hanson, Stephen and Pratt, Lorien},
  journal={Advances in neural information processing systems},
  volume={1},
  year={1988}
}

@article{representations,
  title={Shallowing deep networks: Layer-wise pruning based on feature representations},
  author={Chen, Shi and Zhao, Qi},
  journal={IEEE transactions on pattern analysis and machine intelligence},
  volume={41},
  number={12},
  pages={3048--3056},
  year={2018},
  publisher={IEEE}
}

@article{discriminative,
  title={Discriminative layer pruning for convolutional neural networks},
  author={Jordao, Artur and Lie, Maiko and Schwartz, William Robson},
  journal={IEEE Journal of Selected Topics in Signal Processing},
  volume={14},
  number={4},
  pages={828--837},
  year={2020},
  publisher={IEEE}
}

@inproceedings{question,
  title={To filter prune, or to layer prune, that is the question},
  author={Elkerdawy, Sara and Elhoushi, Mostafa and Singh, Abhineet and Zhang, Hong and Ray, Nilanjan},
  booktitle={Proceedings of the Asian Conference on Computer Vision},
  year={2020}
}

@inproceedings{gal,
  title={Towards optimal structured cnn pruning via generative adversarial learning},
  author={Lin, Shaohui and Ji, Rongrong and Yan, Chenqian and Zhang, Baochang and Cao, Liujuan and Ye, Qixiang and Huang, Feiyue and Doermann, David},
  booktitle={Proceedings of the IEEE/CVF Conference on Computer Vision and Pattern Recognition},
  pages={2790--2799},
  year={2019}
}

@article{svcca,
  title={Svcca: Singular vector canonical correlation analysis for deep learning dynamics and interpretability},
  author={Raghu, Maithra and Gilmer, Justin and Yosinski, Jason and Sohl-Dickstein, Jascha},
  journal={Advances in neural information processing systems},
  volume={30},
  year={2017}
}

@inproceedings{cca,
  title={Deep canonical correlation analysis},
  author={Andrew, Galen and Arora, Raman and Bilmes, Jeff and Livescu, Karen},
  booktitle={International conference on machine learning},
  pages={1247--1255},
  year={2013},
  organization={PMLR}
}

@article{content,
  title={Content and cluster analysis: assessing representational similarity in neural systems},
  author={Laakso, Aarre and Cottrell, Garrison},
  journal={Philosophical psychology},
  volume={13},
  number={1},
  pages={47--76},
  year={2000}
}

@article{extent,
  title={Towards understanding learning representations: To what extent do different neural networks learn the same representation},
  author={Wang, Liwei and Hu, Lunjia and Gu, Jiayuan and Hu, Zhiqiang and Wu, Yue and He, Kun and Hopcroft, John},
  journal={Advances in neural information processing systems},
  volume={31},
  year={2018}
}

@article{shallower,
  title={Layer pruning for obtaining shallower resnets},
  author={Zhang, Ke and Liu, Guangzhe},
  journal={IEEE Signal Processing Letters},
  volume={29},
  pages={1172--1176},
  year={2022},
  publisher={IEEE}
}

@inproceedings{information,
  title={On the information of feature maps and pruning of deep neural networks},
  author={Soltani, Mohammadreza and Wu, Suya and Ding, Jie and Ravier, Robert and Tarokh, Vahid},
  booktitle={2020 25th International Conference on Pattern Recognition (ICPR)},
  pages={6988--6995},
  year={2021},
  organization={IEEE}
}

@article{fisher-segmentation,
  title={The use of multiple measurements in taxonomic problems},
  author={Fisher, Ronald A},
  journal={Annals of eugenics},
  volume={7},
  number={2},
  pages={179--188},
  year={1936},
  publisher={Wiley Online Library}
}

@inproceedings{gradnorm,
  title={Gradnorm: Gradient normalization for adaptive loss balancing in deep multitask networks},
  author={Chen, Zhao and Badrinarayanan, Vijay and Lee, Chen-Yu and Rabinovich, Andrew},
  booktitle={International conference on machine learning},
  pages={794--803},
  year={2018},
  organization={PMLR}
}

@inproceedings{naswot,
  title={Neural architecture search without training},
  author={Mellor, Joe and Turner, Jack and Storkey, Amos and Crowley, Elliot J},
  booktitle={International Conference on Machine Learning},
  pages={7588--7598},
  year={2021},
  organization={PMLR}
}

@article{imagenet,
author = {Russakovsky, Olga and Deng, Jia and Su, Hao and Krause, Jonathan and Satheesh, Sanjeev and Ma, Sean and Huang, Zhiheng and Karpathy, Andrej and Khosla, Aditya and Bernstein, Michael and Berg, Alexander and Fei-Fei, Li},
year = {2014},
month = {09},
pages = {},
title = {ImageNet Large Scale Visual Recognition Challenge},
volume = {115},
journal = {International Journal of Computer Vision},
doi = {10.1007/s11263-015-0816-y}
}

@article{fontana2024distilled,
  title={Distilled Gradual Pruning with Pruned Fine-tuning},
  author={Fontana, Federico and Lanzino, Romeo and Marini, Marco Raoul and Avola, Danilo and Cinque, Luigi and Scarcello, Francesco and Foresti, Gian Luca},
  journal={IEEE Transactions on Artificial Intelligence},
  volume={1},
  number={01},
  pages={1--11},
  year={2024},
  publisher={IEEE Computer Society}
}

@article{chip,
  title={Chip: Channel independence-based pruning for compact neural networks},
  author={Sui, Yang and Yin, Miao and Xie, Yi and Phan, Huy and Aliari Zonouz, Saman and Yuan, Bo},
  journal={Advances in Neural Information Processing Systems},
  volume={34},
  pages={24604--24616},
  year={2021}
}

@inproceedings{elkerdawy2020one,
  title={One-shot layer-wise accuracy approximation for layer pruning},
  author={Elkerdawy, Sara and Elhoushi, Mostafa and Singh, Abhineet and Zhang, Hong and Ray, Nilanjan},
  booktitle={2020 IEEE International Conference on Image Processing (ICIP)},
  pages={2940--2944},
  year={2020},
  organization={IEEE}
}

@inproceedings{o2016radio,
  title={Radio machine learning dataset generation with gnu radio},
  author={O'shea, Timothy J and West, Nathan},
  booktitle={Proceedings of the GNU radio conference},
  volume={1},
  number={1},
  year={2016}
}

@article{chen2021signet,
  title={SigNet: A novel deep learning framework for radio signal classification},
  author={Chen, Zhuangzhi and Cui, Hui and Xiang, Jingyang and Qiu, Kunfeng and Huang, Liang and Zheng, Shilian and Chen, Shichuan and Xuan, Qi and Yang, Xiaoniu},
  journal={IEEE Transactions on Cognitive Communications and Networking},
  volume={8},
  number={2},
  pages={529--541},
  year={2021},
  publisher={IEEE}
}

@article{o2018over,
  title={Over-the-air deep learning based radio signal classification},
  author={O’Shea, Timothy James and Roy, Tamoghna and Clancy, T Charles},
  journal={IEEE Journal of Selected Topics in Signal Processing},
  volume={12},
  number={1},
  pages={168--179},
  year={2018},
  publisher={IEEE}
}

@inproceedings{ganjdanesh2024jointly,
  title={Jointly training and pruning cnns via learnable agent guidance and alignment},
  author={Ganjdanesh, Alireza and Gao, Shangqian and Huang, Heng},
  booktitle={Proceedings of the IEEE/CVF Conference on Computer Vision and Pattern Recognition},
  pages={16058--16069},
  year={2024}
}

@inproceedings{gao2024bilevelpruning,
  title={BilevelPruning: unified dynamic and static channel pruning for convolutional neural networks},
  author={Gao, Shangqian and Zhang, Yanfu and Huang, Feihu and Huang, Heng},
  booktitle={Proceedings of the IEEE/CVF conference on computer vision and pattern recognition},
  pages={16090--16100},
  year={2024}
}

@inproceedings{lu2024entropy,
  title={Entropy induced pruning framework for convolutional neural networks},
  author={Lu, Yiheng and Guan, Ziyu and Yang, Yaming and Zhao, Wei and Gong, Maoguo and Xu, Cai},
  booktitle={Proceedings of the AAAI Conference on Artificial Intelligence},
  volume={38},
  number={4},
  pages={3918--3926},
  year={2024}
}

@article{hossain2024novel,
  title={A Novel Attention-Based Layer Pruning Approach for Low-Complexity Convolutional Neural Networks},
  author={Hossain, Md Bipul and Gong, Na and Shaban, Mohamed},
  journal={Advanced Intelligent Systems},
  volume={6},
  number={11},
  pages={2400161},
  year={2024},
  publisher={Wiley Online Library}
}

\end{document}